\pdfoutput=1
\PassOptionsToPackage{headheight=48pt}{geometry}
\documentclass[]{TEAI}

\usepackage{helvet}
\usepackage[utf8]{inputenc}
\usepackage[T1]{fontenc}
\usepackage{array}
\usepackage{float}
\usepackage{wrapfig}
\usepackage{needspace}
\usepackage{url}
\usepackage{xurl}
\usepackage{tabularx}
\usepackage{colortbl}
\renewcommand{\cfttoctitlefont}{\sectionfont\bfseries\color{seedblue}}

\definecolor{cvprblue}{rgb}{0.21,0.49,0.74}
\definecolor{tabours}{RGB}{237,242,249}
\definecolor{tabno}{RGB}{175,175,175}
\definecolor{tabref}{RGB}{243,243,241}
\newcommand{\cmark}{\textcolor{cvprblue}{$\checkmark$}}
\newcommand{\xmark}{\textcolor{tabno}{$\times$}}
\title{PhyCoVE: Benchmarking Physical Counterfactual Video Editing across Diverse Material Dynamics}

\renewcommand{\authorfont}{\fontsize{11}{13}\selectfont}
\makeatletter
\renewcommand\addtolist[5][]{%
  \begingroup
    \stepcounter{authorcount}%
    \if\relax#3\relax
      \def\sep{}%
    \else
      \ifnum\value{authorcount}=7
        \def\sep{\\[0.8mm]}%
      \else
        \def\sep{#5}%
      \fi
    \fi
    \let\protect\@unexpandable@protect
    \xdef#3{\expandafter{#3}\sep #4[#1]{#2}}%
  \endgroup
}
\makeatother
\renewcommand\author[2][]{\addtolist[#1]{#2}{\authorlist}{\authorformat}{\quad}}
\author[*]{Conghan Yue}
\author[*]{Yuanjie Chen}
\author{Yunyan Xiao}
\author{Zeyu Peng}
\author{Xingsong Ye}
\author{Ya Gao}
\author{Yanglin Huang}
\author{WeiYao Zhang}
\author{Yue Han}
\author[\dagger]{Zhineng Chen}
\author[\dagger]{Yu-Gang Jiang}

\affiliation{Institute of Trustworthy Embodied AI, Fudan University}
\renewcommand{\checkdatafont}{\fontsize{9.5}{13}\selectfont}
\renewcommand\checkdataformat[2][]{{\checkdatafont {\bfseries #1:} #2}}
\checkdata[Code]{\url{https://github.com/Hammour-steak/PhyCoVE}}
\checkdata[Website]{\url{https://phycove.github.io}}
\checkdata[Dataset]{\url{https://huggingface.co/datasets/ccmoony/PhyCoVE}}
\abstract{
Physical counterfactual video editing turns a source video and a language-specified physical edit at a given execution frame into a counterfactual video that shows how the scene would evolve after the edit.
Existing evaluations with counterfactual target videos mainly cover object removal or specific rigid-body motions, although real scenes also involve elastic deformation, fluid flow, and interactions among rigid bodies, elastic solids, and fluids.
Collected videos cannot supply counterfactual video pairs for such scenes, because each records only the physical phenomenon that actually occurred, from which the counterfactual video after an edit is hard to infer.
Constructing new pairs is also difficult, as recording rarely keeps the edit as the only difference between the two videos, especially when a physical quantity must change, whereas simulation requires accurate coupled dynamics and realistic rendering.
We introduce PhyCoVE, a benchmark of 533 cases on 110 source videos that covers rigid, elastic, and fluid dynamics in five interaction patterns, with edits that remove an object, change a physical quantity, or insert a new object.
For the simulated cases, a language model proposes scene designs, coding agents build them with solvers suited to each material and physically based rendering and then simulate each edit from the source scene state, and humans check every stage.
For the real cases, we record continuously from one camera position before and after each edit.
We also propose counterfactual fidelity, a metric that applies across materials and their interactions and measures how much of the physical change from the source video to the target video a generated video reproduces.
We evaluate seven open-source and commercial video editing models and find that counterfactual fidelity ranks them nearly as human preference does, but even the best model reproduces only about one seventh of the physical change.
The models change how a scene looks but rarely how it evolves, whereas a physical edit requires understanding the dynamics that it changes.
Physical counterfactual video editing thus remains far from solved, and PhyCoVE provides a foundation for developing and evaluating methods for it.
}

\patchcmd{\mymaketitle}{\contributionlist\par}{}{}{\PackageError{PhyCoVE}{Could not remove empty contribution block}{Check the TEAI contribution block.}}
\patchcmd{\mymaketitle}{\thispagestyle{firststyle}}{}{}{\PackageError{PhyCoVE}{Could not relocate the first-page style}{Check the TEAI title layout.}}
\patchcmd{\mymaketitle}{\vspace*{0.65cm}}{\vspace*{0pt}}{}{\PackageError{PhyCoVE}{Could not remove empty checkdata spacer}{Check the TEAI title layout.}}

\hypersetup{
    pdftitle={PhyCoVE: Benchmarking Physical Counterfactual Video Editing across Diverse Material Dynamics},
    pdfauthor={Conghan Yue, Yuanjie Chen, Yunyan Xiao, Zeyu Peng, Xingsong Ye, Ya Gao, Yanglin Huang, WeiYao Zhang, Yue Han, Zhineng Chen, Yu-Gang Jiang}
}

\begin{document}
\addtocontents{toc}{\protect\setcounter{tocdepth}{-1}}
\thispagestyle{firststyle}
\begingroup
\renewcommand{\thefootnote}{\fnsymbol{footnote}}
\footnotetext[1]{Equal contribution.}
\footnotetext[2]{Corresponding authors.}
\endgroup
\maketitle
\clearpage

\setcounter{topnumber}{4}
\setcounter{bottomnumber}{3}
\setcounter{totalnumber}{6}
\renewcommand{\topfraction}{0.95}
\renewcommand{\bottomfraction}{0.9}
\renewcommand{\textfraction}{0.05}
\renewcommand{\floatpagefraction}{0.95}
\setlength{\textfloatsep}{8pt plus 2pt minus 2pt}
\setlength{\intextsep}{6pt plus 1pt minus 1pt}
\setlength{\floatsep}{6pt plus 2pt minus 2pt}
\captionsetup[figure]{skip=6pt}
\captionsetup[table]{skip=6pt}
\makeatletter
\setlength{\@fptop}{0pt}
\setlength{\@fpsep}{8pt plus 1pt minus 1pt}
\setlength{\@fpbot}{0pt plus 1fil}
\makeatother
\raggedbottom

\begingroup
\makeatletter
\@clubpenalty=10000
\makeatother
\clubpenalty=10000
\section{Introduction}
\label{sec:introduction}

\begin{figure}[t]
\centering
\includegraphics[width=\linewidth]{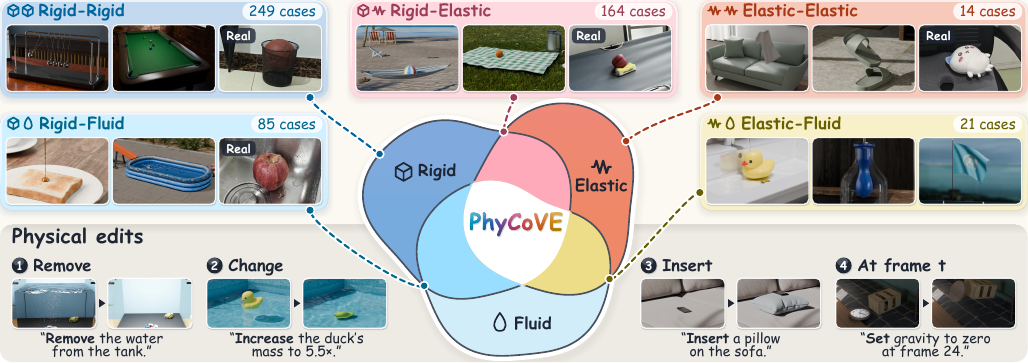}
\caption{\textbf{Overview of PhyCoVE.} The three shapes in the center stand for the three materials, an overlap of two shapes marks the interaction pattern of those materials, and the rigid-rigid and elastic-elastic patterns lie in single shapes. The panels show example scenes and case counts of the five patterns, and the bottom row shows the three edit types and an edit at a later execution frame.}
\label{fig:overview}
\end{figure}

Physical counterfactual video editing~\cite{Yue2026VideoPhysEdit} takes a source video as the condition and a language-specified physical edit applied at a given execution frame as the intervention to generate the corresponding counterfactual video.
The generated video must preserve the factual history before the intervention and afterward depict the physical evolution induced by the edit.
This includes both changes in the edited object's own motion, deformation, or flow and effects on other parts of the scene.

Evaluating model performance on this task requires assessing whether the generated video accurately depicts how the source scene would subsequently evolve under the specified physical intervention, and counterfactual target videos paired with their source videos can provide a direct reference.
However, existing evaluations remain limited in coverage.
VOID~\cite{Motamed2026VOID} focuses on object removal, whereas PCVE-RigidBench~\cite{Yue2026VideoPhysEdit}, despite supporting various physical edits, mainly covers specific rigid-body motions and scene interactions.
In fact, real-world scenes also commonly involve elastic deformation, fluid flow, and interactions among rigid bodies, elastic solids, and fluids.
Evaluation should therefore cover diverse material dynamics to better assess models' editing capabilities in complex physical scenes.

Constructing counterfactual video pairs for such scenes is challenging.
In real-world scenes, a single recording captures only one course of physical evolution, while numerous uncontrollable factors make it difficult to precisely reproduce the original process and thus obtain counterfactual videos whose differences arise solely from the specified physical edit.
Physical quantities are even harder to obtain and control, because changing one requires an object that looks the same but differs only in that quantity by an exact amount, such as a ping-pong ball exactly ten times heavier, and gravity is harder still to edit.
In simulated scenes, every quantity can be set to an exact value, and the counterfactual simulation can start from the same scene state at the execution frame.
However, scenes with interacting materials are difficult to build in simulation, because their material properties and interaction conditions must be configured carefully, their coupled dynamics must be simulated accurately, and the scenes must be rendered with high visual realism.
Therefore, in real-world scenes it is hard to keep the edit as the only difference between the two videos, whereas in simulated scenes the coupled dynamics must be accurate and the rendering realistic.

\begin{figure}[t]
\centering
\includegraphics[width=\linewidth]{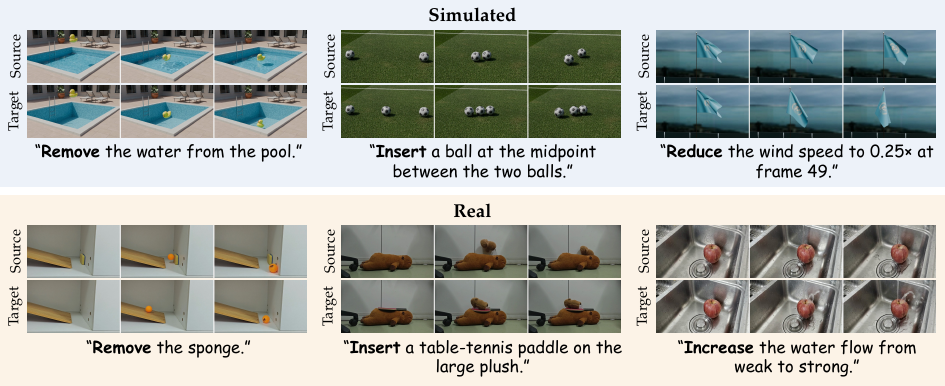}
\caption{\textbf{Representative cases from PhyCoVE.} Each case pairs a source video and a physical edit at its execution frame with a counterfactual target video.}
\label{fig:cases}
\end{figure}

We introduce \textbf{PhyCoVE}, a \textbf{Phy}sical \textbf{Co}unterfactual \textbf{V}ideo \textbf{E}diting benchmark that covers rigid, elastic, and fluid dynamics and their interactions in a simulated set and a real set.
For the simulated set, a language model proposes the scene designs, coding agents build and iteratively refine the scenes in simulators, and humans check the results after each stage.
Starting from the source scene state at the execution frame, we simulate each edit with solvers suited to the materials and render both videos with the same assets, camera, and lighting through physically based rendering, so their differences arise only from the edit.
For the real set, we record each scene continuously from one camera position, cut takes, align them in time, and match their exposure.
PhyCoVE contains 533 \emph{cases} in 110 \emph{source groups}, where each case pairs a source video and a physical edit with a counterfactual target video, and the cases in a source group share one source video.
Figure~\ref{fig:overview} shows the interaction patterns and edits of PhyCoVE, and Figure~\ref{fig:cases} shows cases from both sets.
Measured by counterfactual fidelity, which compares each generated video with the source and target videos of its case, even the best of seven open-source and commercial video editing models reproduces only about one seventh of the physical change that an edit causes.
The models change how a scene looks but rarely how it evolves, whereas a physical edit requires understanding the dynamics that it changes.
Physical counterfactual video editing thus remains far from solved, and PhyCoVE provides a foundation for developing and evaluating methods for it.

Our main contributions are as follows:
\begin{itemize}
    \item We introduce PhyCoVE, which contains 533 human-checked cases in 110 source groups. The cases cover rigid, elastic, and fluid dynamics in five interaction patterns, with edits that remove an object, change a physical quantity, or insert a new object, some of which take effect at a later frame, and the 508 simulated cases also provide per-frame states of their rigid and elastic objects.
    \item We propose an evaluation protocol that uses counterfactual fidelity to score how far a generated video follows the physical edit across rigid, elastic, and fluid responses and keeps the source video before the execution frame, and a blind human study supports its ranking of the models.
    \item We evaluate seven video editing models and find that they mostly remove only the appearance of an object and often do so before the execution frame, hardly follow changes of a physical quantity, and rank differently across materials. These findings show that current models change how a scene looks but rarely how it evolves, and that physical counterfactual video editing remains far from solved.
\end{itemize}

\endgroup
\section{Related Work}
\label{sec:related_work}

\begin{table}[!htbp]
\centering
\caption{\textbf{Comparison with related benchmarks.} Task is video editing (Edit), video generation (Gen.), question answering (QA), or state prediction (Pred.). Target, States, and Real mark counterfactual target videos, per-frame object states, and real videos, and R, E, and F stand for rigid, elastic or deformable, and fluid dynamics. $^{\dagger}$Only for synthetic test videos.}
\label{tab:comparison}
\small
\setlength{\tabcolsep}{6pt}
\begin{tabularx}{0.88\linewidth}{ll>{\centering\arraybackslash\hsize=1.25\hsize}X>{\centering\arraybackslash\hsize=1.25\hsize}X*{3}{>{\centering\arraybackslash\hsize=0.8\hsize}X}>{\centering\arraybackslash\hsize=1.1\hsize}X}
\toprule
\multirow{2}{*}[-0.55ex]{\textbf{Benchmark}} & \multirow{2}{*}[-0.55ex]{\textbf{Task}} & \multicolumn{2}{c}{\textbf{Reference}} & \multicolumn{3}{c}{\textbf{Dynamics}} & \multirow{2}{*}[-0.55ex]{\textbf{Real}} \\
\cmidrule(lr){3-4}\cmidrule(lr){5-7}
 & & \textbf{Target} & \textbf{States} & \textbf{R} & \textbf{E} & \textbf{F} & \\
\midrule
NRVBench~\cite{Qu2026BeyondRigid} & Edit & \xmark & \xmark & \xmark & \cmark & \cmark & \cmark \\
VOID~\cite{Motamed2026VOID} & Edit & \cmark$^{\dagger}$ & \xmark & \cmark & \cmark & \xmark & \cmark \\
PCVE-RigidBench~\cite{Yue2026VideoPhysEdit} & Edit & \cmark & \cmark & \cmark & \xmark & \xmark & \xmark \\
\midrule
VideoPhy~\cite{Bansal2025VideoPhy} & Gen. & \xmark & \xmark & \cmark & \cmark & \cmark & \xmark \\
Physics-IQ~\cite{Motamed2026PhysicsIQ} & Gen. & \xmark & \xmark & \cmark & \cmark & \cmark & \cmark \\
CRONOS~\cite{Begiristain2026CRONOS} & Gen. & \xmark & \xmark & \cmark & \xmark & \xmark & \xmark \\
\midrule
Physion++~\cite{Tung2023PhysionPlusPlus} & QA & \xmark & \cmark & \cmark & \cmark & \cmark & \xmark \\
ContPhy~\cite{Zheng2024ContPhy} & QA & \xmark & \cmark & \cmark & \cmark & \cmark & \xmark \\
CoPhy~\cite{Baradel2020CoPhy} & Pred. & \cmark & \cmark & \cmark & \xmark & \xmark & \xmark \\
\midrule
\rowcolor{tabours}\textbf{PhyCoVE} & Edit & \cmark & \cmark & \cmark & \cmark & \cmark & \cmark \\
\bottomrule
\end{tabularx}
\end{table}

Table~\ref{tab:comparison} compares PhyCoVE with related benchmarks.

\textbf{Video editing benchmarks.}
Video editing benchmarks~\cite{Wu2023TGVE,Chen2025EditBoard,Sun2025VEBench} evaluate whether an edited video follows the text prompt and preserves the rest of the source video.
FiVE-Bench~\cite{Li2025FiVEBench} covers appearance and object edits, IVEBench~\cite{Chen2026IVEBench} also covers motion and camera edits, and both score the outputs against the text prompt and the source video with standard metrics and vision-language models.
NRVBench~\cite{Qu2026BeyondRigid} focuses on non-rigid edits of soft bodies, cloth, hair, liquids, and smoke in real videos and uses vision-language models to judge their physical plausibility.
These benchmarks provide no target video, whereas VOID~\cite{Motamed2026VOID} removes objects together with the interactions they cause and has counterfactual target videos for its synthetic test cases.
VideoPhysEdit~\cite{Yue2026VideoPhysEdit} builds PCVE-RigidBench, which pairs each source video and physical edit with a counterfactual target video and object states in rigid-body scenes.
Across rigid, elastic, and fluid dynamics and their interactions, PhyCoVE pairs physical edits with counterfactual target videos and, in its simulated set, the states of rigid and elastic objects.

\textbf{Physics in video generation.}
Physics benchmarks for video generation~\cite{Li2025PISA,Tragoudaras2026Morpheus,Li2025WorldModelBench,Zheng2025VBench2} assess whether generated videos follow physical laws.
VideoPhy~\cite{Bansal2025VideoPhy} organizes text prompts by solid-solid, solid-fluid, and fluid-fluid interactions, VideoPhy-2~\cite{Bansal2026VideoPhy2} focuses on physical commonsense in real-world actions, and PhyGenBench~\cite{Meng2025PhyGenBench} covers 27 physical laws.
All three rate text-to-video outputs with humans or vision-language models, without reference videos.
Physics-IQ~\cite{Motamed2026PhysicsIQ} instead compares generated continuations of recorded experiments with the recorded future frames.
CRONOS~\cite{Begiristain2026CRONOS} and PhysEditWorld~\cite{Hu2026PhysEditWorld} construct matched videos under controlled changes, where CRONOS varies the viewpoint, scene, object category, and appearance of rigid-body events and PhysEditWorld replays the same game actions under different gravity settings.
PhysInOne~\cite{Zhou2026PhysInOne} simulates rigid, deformable, granular, and liquid phenomena for physics-aware video generation~\cite{Liu2024PhysGen,Wang2025PhysCtrl,Xue2025PhyT2V,Wang2025WISA}.
In these works, models generate videos from text or observed frames, whereas PhyCoVE asks models to apply a physical edit to a source video and evaluates each result against its counterfactual target video.

\textbf{Physical and counterfactual reasoning.}
Reasoning benchmarks~\cite{Riochet2022IntPhys,Bear2021Physion,Chen2022ComPhy,Ates2022CRAFT,Chow2025PhysBench} probe physical understanding through answers or predicted object states rather than generated videos.
Physion++~\cite{Tung2023PhysionPlusPlus} asks models to predict object contact in scenes whose outcomes depend on mass, friction, elasticity, or deformability, including scenes with liquids and cloth.
ContPhy~\cite{Zheng2024ContPhy} asks property and dynamics questions, including counterfactual ones, about liquids, cloth, ropes, and elastoplastic balls, while CLEVRER~\cite{Yi2020CLEVRER} and Causal-VidQA~\cite{Li2022CausalVidQA} ask counterfactual questions about rigid collisions and human actions, respectively.
CoPhy~\cite{Baradel2020CoPhy} moves or removes objects in the initial frame and asks models to predict the resulting poses of rigid objects.
Physion++, ContPhy, CLEVRER, and CoPhy build their scenes in simulators and record object states behind each outcome.
PhyCoVE likewise records object states in simulation but asks models to generate the counterfactual video.

\section{The PhyCoVE Benchmark}
\label{sec:benchmark}

\subsection{Task and Data Format}
\label{sec:task}
In physical counterfactual video editing, a model receives a source video and a language instruction that specifies a physical edit and its execution frame, and generates the counterfactual video.
Each case in PhyCoVE consists of a source video, an instruction, and a counterfactual target video, and each simulated case also provides the states of its rigid and elastic objects at every frame.
All videos have 96 frames at 24 fps with a resolution of $1280\times720$.
In the simulated set, every instruction follows a template that names the edit, the edited object or the scene, and the execution frame, as in ``Increase the yellow duck's mass to 5.5 times its original value from frame 1 onwards.''
Such an instruction states only the edit, so the model must infer the resulting motion, deformation, or flow on its own.

\begin{figure}[!htbp]
\centering
\includegraphics[width=\linewidth]{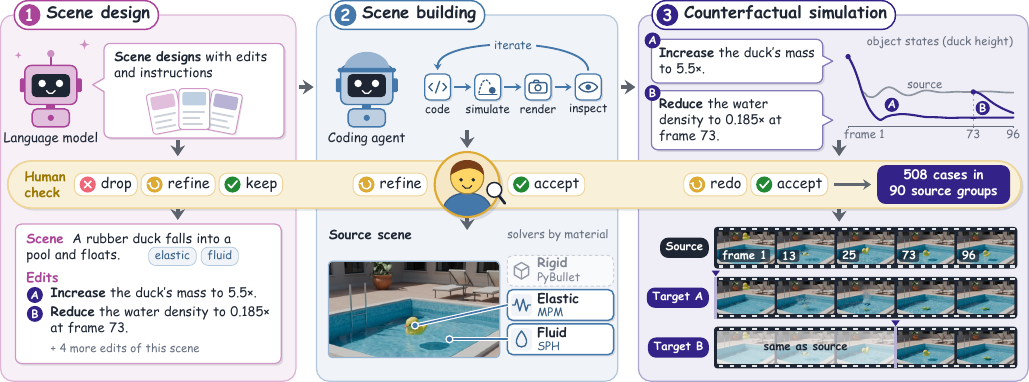}
\caption{\textbf{Construction of the simulated set.} A language model proposes scene designs, coding agents build the selected designs into source scenes, we simulate each edit from its execution frame, and humans check the result of every stage.}
\label{fig:pipeline}
\end{figure}

\subsection{Scenes and Edits}
\label{sec:scenes}
Figure~\ref{fig:overview} summarizes the scenes and edits.
PhyCoVE contains 110 scenes, one for each source group, and we assign each scene to one of five interaction patterns, namely rigid-rigid, rigid-elastic, rigid-fluid, elastic-fluid, and elastic-elastic, according to the materials that interact.
Each case applies one physical edit, whose type is a removal of an object or a fluid, a change of a physical quantity, or an insertion of a new object or a fluid.
A change scales or sets a quantity such as the mass, friction, restitution, stiffness, density, or viscosity of an object or a fluid, the initial velocity of an object, the flow rate of a faucet, the wind speed, or the gravity of the scene.
Each edit also has an execution frame, which is the first frame in 470 cases and a later frame in 63 cases, and the generated video must match the source video before that frame.

\subsection{Construction of the Simulated Set}
\label{sec:construction}
Figure~\ref{fig:pipeline} shows the three stages in which we construct the simulated set, and every stage ends with a human check.

\textbf{Scene design.}
A language model proposes scene designs as in~\cite{Yang2024Holodeck,Hu2024SceneCraft}, each consisting of a scene description and physical edits with their instructions, and humans select the designs to build or ask the model to refine them.

\textbf{Scene building.}
Coding agents implement each selected design as a program that simulates the source scene and produces its video with physically based rendering in Blender~\cite{Blender2024}.
The program simulates each material with a solver suited to it, for example PyBullet~\cite{Coumans2016PyBullet} or MuJoCo~\cite{Todorov2012MuJoCo} for rigid objects, incremental potential contact (IPC)~\cite{Li2020IPC} or the material point method (MPM)~\cite{Hu2018MLSMPM} for elastic objects, and the smoothed-particle hydrodynamics (SPH) solver~\cite{Monaghan1992SPH} of Genesis~\cite{Genesis2024Genesis} or the fluid-implicit-particle (FLIP) solver~\cite{Zhu2005Sand} of Blender for liquids.
Humans then check whether the source video shows physically plausible motion and realistic rendering, and accept the scene or ask the agents to refine it.

\textbf{Counterfactual simulation.}
For every edit of an accepted scene, we start from the source scene state at the execution frame, apply the edit, and simulate the rest of the video.
Rendering the result with the assets, camera, and lighting of the source video ensures that the differences between the two videos arise only from the edit.
The simulation also exports the states of the rigid and elastic objects, and in every case with a later execution frame, these states are identical to those of the source scene before that frame.
Humans then check whether each case shows the specified edit and a physically plausible response, and accept it or have it redone.
The 508 accepted cases in 90 source groups form the simulated set.

\subsection{Construction of the Real Set}
\label{sec:real}
For each source group of the real set, we record continuously from one camera position, first the source scene and then each edit that we set up by hand, for example by removing or inserting an object.
We cut this recording into takes, align them in time, correct exposure differences between them, and convert them to the shared video format.
The take of the source scene serves as the source video and the take after an edit as the counterfactual target video, and together with a manually written instruction they form a real case.
Humans then check whether each real case shows the specified edit.
The 25 cases in 20 source groups, whose scenes involve rigid objects, elastic objects such as plush toys and cushions, and water, form the real set.

\Needspace{10\baselineskip}
\section{Experiments}
\label{sec:experiments}

\begin{table}[!htbp]
\centering
\caption{\textbf{Results on PhyCoVE.} HPWR is the human preference win rate in percent. The visual fidelity metrics compare generated videos with target videos, No edit scores the unedited source video, and the last two rows give the Kendall $\tau$~\cite{Kendall1938Tau} and the Spearman $\rho$~\cite{Spearman1904Rho} of each metric with HPWR over the seven models. Bold and underline mark the best and second best among the models.}
\label{tab:main}
\footnotesize
\setlength{\tabcolsep}{2.5pt}
\begin{tabularx}{\linewidth}{l>{\centering\arraybackslash\hsize=0.95\hsize}X>{\centering\arraybackslash}X>{\centering\arraybackslash\hsize=0.95\hsize}X>{\centering\arraybackslash\hsize=0.95\hsize}X>{\centering\arraybackslash\hsize=0.95\hsize}X>{\centering\arraybackslash\hsize=1.2\hsize}X>{\centering\arraybackslash}X>{\centering\arraybackslash}X>{\centering\arraybackslash}X>{\centering\arraybackslash}X}
\toprule
\multirow{2}{*}[-0.55ex]{\textbf{Model}} & \textbf{Human} & \multicolumn{5}{c}{\textbf{Physical metrics}} & \multicolumn{4}{c}{\textbf{Visual fidelity}} \\
\cmidrule(lr){2-2}\cmidrule(lr){3-7}\cmidrule(lr){8-11}
 & \textbf{HPWR}\,$\uparrow$ & \textbf{CF}\,$\uparrow$ & \textbf{EA}\,$\uparrow$ & \textbf{PC}\,$\uparrow$ & \textbf{Joint}\,$\uparrow$ & \mbox{\textbf{Mask IoU}\kern1pt$\uparrow$} & \textbf{PSNR}\,$\uparrow$ & \textbf{SSIM}\,$\uparrow$ & \textbf{LPIPS}\,$\downarrow$ & \textbf{FVD}\,$\downarrow$ \\
\midrule
\multicolumn{11}{l}{\textit{Open-source}} \\
JoyAI~\cite{Xiao2026JoyAIVideoEdit} & 44.9 & 0.100 & 1.86 & 4.15 & 7.7 & \underline{0.331} & 24.28 & 0.792 & 0.196 & 257 \\
Kiwi-Edit~\cite{Lin2026KiwiEdit} & 44.1 & $-$0.005 & 1.86 & 4.03 & 6.0 & 0.325 & \underline{26.30} & \underline{0.860} & \underline{0.113} & 131 \\
OmniWeaving~\cite{Pan2026OmniWeaving} & 35.8 & $-$0.182 & 1.56 & 4.05 & 2.8 & 0.222 & 22.17 & 0.813 & 0.274 & 282 \\
\midrule
\multicolumn{11}{l}{\textit{Commercial}} \\
Kling 4.0~\cite{Kuaishou2026Kling} & 51.6 & 0.107 & 1.88 & \underline{4.41} & 8.3 & \textbf{0.334} & 23.48 & 0.806 & 0.127 & 135 \\
MiniMax H3~\cite{MiniMax2026H3} & \underline{59.0} & \textbf{0.134} & \textbf{2.20} & 4.39 & \underline{11.1} & 0.306 & 23.44 & 0.785 & 0.155 & \underline{105} \\
Seedance 2.5~\cite{ByteDance2026Seedance} & \textbf{59.4} & \underline{0.126} & \underline{2.07} & \textbf{4.42} & \textbf{12.0} & 0.330 & \textbf{28.28} & \textbf{0.891} & \textbf{0.086} & \textbf{87} \\
Wan 3.0~\cite{Alibaba2026Wan} & 55.3 & 0.113 & 1.99 & \underline{4.41} & 9.9 & 0.327 & 26.15 & 0.843 & 0.121 & 166 \\
\midrule
\rowcolor{tabref}No edit & -- & 0.000 & -- & -- & -- & 0.359 & 30.03 & 0.960 & 0.043 & 147 \\
\midrule
$\tau$ with HPWR & -- & 0.90 & 0.90 & 0.62 & 1.00 & 0.24 & 0.24 & 0.05 & 0.43 & 0.62 \\
$\rho$ with HPWR & -- & 0.96 & 0.96 & 0.82 & 1.00 & 0.29 & 0.39 & 0.07 & 0.54 & 0.75 \\
\bottomrule
\end{tabularx}
\end{table}

\subsection{Models and Metrics}
\label{sec:metrics}
We evaluate the open-source JoyAI-Video-Edit~\cite{Xiao2026JoyAIVideoEdit} (JoyAI), Kiwi-Edit~\cite{Lin2026KiwiEdit}, and OmniWeaving~\cite{Pan2026OmniWeaving}, and the commercial Kling 4.0~\cite{Kuaishou2026Kling}, MiniMax H3~\cite{MiniMax2026H3}, Seedance 2.5~\cite{ByteDance2026Seedance}, and Wan 3.0~\cite{Alibaba2026Wan}.
Each model receives the source video and the instruction of a case and generates one video.
We also evaluate VOID~\cite{Motamed2026VOID} on the removal cases and VideoPhysEdit~\cite{Yue2026VideoPhysEdit} on cases with classic rigid-body motion.
Appendices~\ref{sec:supp_cf} and~\ref{sec:supp_judge} give the full protocol of the metrics below.

\begin{figure}[!htbp]
\centering
\includegraphics[width=0.58\linewidth]{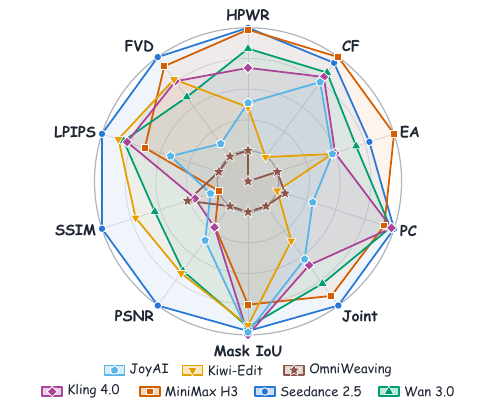}
\caption{\textbf{Results of the seven models on the metrics of Table~\ref{tab:main}.} Each axis runs from the lowest value among the models on the inner ring to the highest on the outer ring, with LPIPS and FVD reversed, but the inner ring of the CF axis marks a CF of 0.}
\label{fig:radar}
\end{figure}

\textbf{Counterfactual fidelity.}
Counterfactual fidelity (CF) measures how much of the physical change from the source video to the target video a generated video reproduces.
We locate every object of a case from its recorded states in the simulated cases and from annotations in the real cases, and track its image region through each video with SAM~2~\cite{Ravi2025SAM2} and CoTracker3~\cite{Karaev2024CoTracker3}.
From each track we measure quantities of the object's motion, such as its final position, the length of its path, and the time at which it comes to rest.
For an elastic object we also measure its height, width, and area, for a fluid its amount and extent over time, and for a removed or inserted object its presence.
For a measured quantity with value $q^{\mathrm{src}}$ in the source video, $q^{\mathrm{tgt}}$ in the target video, and $q^{\mathrm{gen}}$ in the generated video, the score is
\begin{equation}
\mathrm{CF}(q)=\frac{\langle q^{\mathrm{gen}}-q^{\mathrm{src}},\,q^{\mathrm{tgt}}-q^{\mathrm{src}}\rangle}{\lVert q^{\mathrm{tgt}}-q^{\mathrm{src}}\rVert^{2}}.
\label{eq:cf}
\end{equation}
As in the physical edit score that VideoPhysEdit~\cite{Yue2026VideoPhysEdit} computes for rigid objects, a score of 1 means that the generated video reproduces the whole change, a score of 0 that it keeps the behavior of the source video, and a negative score that it moves the quantity away from the target.
We subtract the excess from the score of a change that overshoots the target, assign a score of $-1$ to every quantity of an object that vanishes from the generated video but remains in the target video, and clip all scores to $[-1,1]$.
The CF of a video is the mean over its scored quantities, which are those that differ clearly between the source and target videos, and the CF of a model is the mean over its videos.
For an edit at a later frame, the target video equals the source video before the execution frame, so we also subtract from the scores of each object how far it departs from the source video over these frames, relative to how far the target video departs over the frames that follow.
Because objects in the generated videos can deform, duplicate, or vanish, humans verify every track against its video.
Under this protocol, a re-encoded copy of the target video obtains a CF of 0.960, whereas a re-encoded copy of the source video obtains 0.004.

\textbf{Judge scores.}
Following VideoPhy-2~\cite{Bansal2026VideoPhy2}, we use GPT-6 Astra~\cite{OpenAI2026GPT6Astra} as a vision-language judge~\cite{Zheng2023MTBench,Chen2024MLLMJudge} that gives each generated video two scores from 1 to 5.
Edit adherence (EA) measures whether the generated video shows the same change as the target video, with the right object, direction, magnitude, and timing, and the judge sees the instruction and the frames of the source, target, and generated videos.
An EA of 1 means that the generated video keeps the behavior of the source video or that its change has the wrong object or direction.
An EA of 3 means that the change has the right object and direction but one substantial mismatch, and an EA of 5 that the change and its consequences match those of the target video.
Physical commonsense (PC) measures whether the motion, contact, deformation, and flow in the generated video are physically plausible, and the judge sees only the frames of the generated video.
Joint is the percentage of videos with an EA and a PC of at least 4.

\textbf{Mask IoU and visual fidelity.}
As a further physical metric, Mask IoU~\cite{Yue2026VideoPhysEdit} measures how well the object regions of the generated video overlap those of the target video.
For visual fidelity, we report PSNR, SSIM~\cite{Wang2004SSIM}, and LPIPS~\cite{Zhang2018LPIPS} per video pair and FVD~\cite{Unterthiner2018FVD} between the two video sets.

\textbf{Human preference.}
We also run a blind pairwise study, in which annotators see the instruction, the source and target videos, and the anonymous videos of two models for a case, and choose the one that better completes the edit or call a tie.
Annotators judge edit adherence first, then physical plausibility, and then the preservation of the rest of the scene.
Ten annotators judge 500 comparisons that cover every pair of models 23 or 24 times, and three of them judge each comparison independently.
The human preference win rate (HPWR) of a model is its share of wins, with a tie as half a win, so 50\% marks parity.
Appendix~\ref{sec:supp_human} gives the protocol and the full results.

\begin{figure}[!htbp]
\centering
\includegraphics[width=0.80\linewidth]{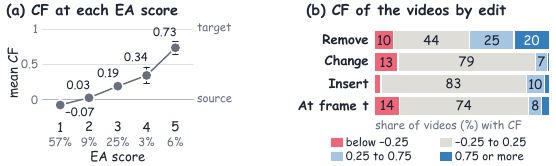}
\caption{\textbf{CF of the videos of the seven models.} (a) Mean CF at each EA with 95\% bootstrap confidence intervals~\cite{Efron1993Bootstrap}, and the percentage of videos at each EA below the axis. (b) Share of videos in each CF range by edit.}
\label{fig:analysis}
\end{figure}

\begin{figure}[!htbp]
\centering
\includegraphics[width=0.80\linewidth]{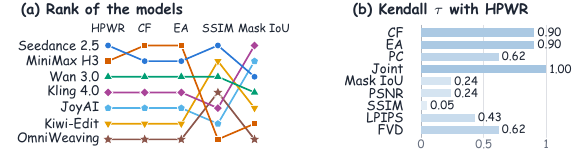}
\caption{\textbf{Agreement with human preference.} (a) Rank of the seven models by HPWR, CF, EA, SSIM, and Mask IoU, with the first at the top. (b) Kendall $\tau$ of each metric of Table~\ref{tab:main} with HPWR.}
\label{fig:agreement}
\end{figure}

\begin{table}[!htbp]
\centering
\caption{\textbf{Results by edit type and interaction pattern.} HPWR comes from the comparisons within each edit type, and R, E, and F mean rigid, elastic, and fluid. Blue and red shades mark values above and below 0, 1 for EA, or 50 for HPWR, and bold marks the column best.}
\label{tab:breakdown}
\footnotesize
\setlength{\tabcolsep}{3pt}
\resizebox{\linewidth}{!}{%
\begin{tabular}{l*{14}{c}}
\toprule
\multirow{3}{*}[-1.1ex]{\textbf{Model}} & \multicolumn{9}{c}{\textbf{Edit type}} & \multicolumn{5}{c}{\textbf{Interaction pattern} (CF)} \\
\cmidrule(lr){2-10}\cmidrule(lr){11-15}
 & \multicolumn{3}{c}{Remove} & \multicolumn{3}{c}{Change} & \multicolumn{3}{c}{Insert} & \multirow{2}{*}[-0.55ex]{R-R} & \multirow{2}{*}[-0.55ex]{R-E} & \multirow{2}{*}[-0.55ex]{R-F} & \multirow{2}{*}[-0.55ex]{E-F} & \multirow{2}{*}[-0.55ex]{E-E} \\
\cmidrule(lr){2-4}\cmidrule(lr){5-7}\cmidrule(lr){8-10}
 & CF & EA & HPWR & CF & EA & HPWR & CF & EA & HPWR & & & & & \\
\midrule
\multicolumn{15}{l}{\textit{Open-source}} \\
JoyAI & \cellcolor[RGB]{190,213,233}0.266 & \cellcolor[RGB]{185,210,232}2.71 & \cellcolor[RGB]{251,220,224}43.1 & \cellcolor[RGB]{253,254,254}0.007 & \cellcolor[RGB]{239,245,250}1.38 & \cellcolor[RGB]{252,229,232}44.9 & \cellcolor[RGB]{230,239,247}0.104 & \cellcolor[RGB]{197,217,236}\textbf{2.43} & \cellcolor[RGB]{185,210,232}\textbf{61.7} & \cellcolor[RGB]{235,242,248}0.084 & \cellcolor[RGB]{230,239,247}0.103 & \cellcolor[RGB]{213,228,241}0.173 & \cellcolor[RGB]{252,253,254}0.011 & \cellcolor[RGB]{234,242,248}0.085 \\
Kiwi-Edit & \cellcolor[RGB]{202,221,237}0.216 & \cellcolor[RGB]{183,209,231}2.76 & \cellcolor[RGB]{253,237,239}46.5 & \cellcolor[RGB]{251,215,219}$-$0.133 & \cellcolor[RGB]{238,244,249}1.40 & \cellcolor[RGB]{251,220,224}43.1 & \cellcolor[RGB]{247,250,252}0.034 & \cellcolor[RGB]{232,240,247}1.57 & \cellcolor[RGB]{250,204,210}40.0 & \cellcolor[RGB]{255,251,251}$-$0.014 & \cellcolor[RGB]{247,250,252}0.031 & \cellcolor[RGB]{255,251,252}$-$0.013 & \cellcolor[RGB]{255,252,253}$-$0.008 & \cellcolor[RGB]{249,198,204}$-$0.186 \\
OmniWeaving & \cellcolor[RGB]{252,229,231}$-$0.087 & \cellcolor[RGB]{210,226,240}2.09 & \cellcolor[RGB]{247,179,187}27.8 & \cellcolor[RGB]{247,180,188}$-$0.247 & \cellcolor[RGB]{243,247,251}1.29 & \cellcolor[RGB]{250,206,211}40.3 & \cellcolor[RGB]{254,247,248}$-$0.026 & \cellcolor[RGB]{234,241,248}1.52 & \cellcolor[RGB]{247,179,187}30.6 & \cellcolor[RGB]{248,186,193}$-$0.227 & \cellcolor[RGB]{250,212,216}$-$0.142 & \cellcolor[RGB]{251,216,220}$-$0.127 & \cellcolor[RGB]{250,204,209}$-$0.169 & \cellcolor[RGB]{250,209,214}$-$0.150 \\
VOID~\cite{Motamed2026VOID} & \cellcolor[RGB]{183,208,231}0.297 & \cellcolor[RGB]{177,205,229}2.90 & -- & -- & -- & -- & -- & -- & -- & -- & -- & -- & -- & -- \\
\midrule
\multicolumn{15}{l}{\textit{Commercial}} \\
Kling 4.0 & \cellcolor[RGB]{185,210,232}0.289 & \cellcolor[RGB]{180,207,230}2.84 & \cellcolor[RGB]{245,248,252}51.7 & \cellcolor[RGB]{255,255,255}0.000 & \cellcolor[RGB]{241,246,250}1.35 & \cellcolor[RGB]{247,250,252}51.3 & \cellcolor[RGB]{213,228,241}\textbf{0.173} & \cellcolor[RGB]{206,224,239}2.19 & \cellcolor[RGB]{215,229,242}56.7 & \cellcolor[RGB]{232,240,247}0.097 & \cellcolor[RGB]{221,233,244}0.140 & \cellcolor[RGB]{231,240,247}0.097 & \cellcolor[RGB]{238,244,249}0.071 & \cellcolor[RGB]{246,249,252}0.037 \\
MiniMax H3 & \cellcolor[RGB]{173,202,228}0.336 & \cellcolor[RGB]{167,198,226}\textbf{3.15} & \cellcolor[RGB]{174,203,228}\textbf{63.5} & \cellcolor[RGB]{250,252,253}\textbf{0.019} & \cellcolor[RGB]{227,237,246}\textbf{1.69} & \cellcolor[RGB]{215,229,242}56.7 & \cellcolor[RGB]{222,234,244}0.136 & \cellcolor[RGB]{200,220,237}2.33 & \cellcolor[RGB]{208,225,239}57.8 & \cellcolor[RGB]{229,238,246}0.107 & \cellcolor[RGB]{213,228,241}\textbf{0.171} & \cellcolor[RGB]{218,231,243}0.154 & \cellcolor[RGB]{207,224,239}\textbf{0.197} & \cellcolor[RGB]{254,249,250}$-$0.019 \\
Seedance 2.5 & \cellcolor[RGB]{169,200,226}\textbf{0.354} & \cellcolor[RGB]{171,201,227}3.05 & \cellcolor[RGB]{203,221,238}58.7 & \cellcolor[RGB]{254,255,255}0.003 & \cellcolor[RGB]{233,241,248}1.55 & \cellcolor[RGB]{194,216,235}\textbf{60.2} & \cellcolor[RGB]{238,244,249}0.070 & \cellcolor[RGB]{212,227,241}2.05 & \cellcolor[RGB]{242,246,251}52.2 & \cellcolor[RGB]{237,243,249}0.075 & \cellcolor[RGB]{214,229,241}0.168 & \cellcolor[RGB]{203,221,238}\textbf{0.215} & \cellcolor[RGB]{247,250,252}0.034 & \cellcolor[RGB]{202,221,237}\textbf{0.216} \\
Wan 3.0 & \cellcolor[RGB]{169,199,226}\textbf{0.354} & \cellcolor[RGB]{169,199,226}3.11 & \cellcolor[RGB]{203,221,238}58.7 & \cellcolor[RGB]{254,250,250}$-$0.017 & \cellcolor[RGB]{239,245,250}1.40 & \cellcolor[RGB]{232,240,247}53.8 & \cellcolor[RGB]{250,252,253}0.019 & \cellcolor[RGB]{212,227,241}2.05 & \cellcolor[RGB]{248,251,253}51.1 & \cellcolor[RGB]{226,236,245}\textbf{0.120} & \cellcolor[RGB]{226,236,245}0.120 & \cellcolor[RGB]{234,242,248}0.085 & \cellcolor[RGB]{218,231,243}0.152 & \cellcolor[RGB]{253,253,254}0.010 \\
\bottomrule
\end{tabular}}
\end{table}

\begin{figure}[!htbp]
\centering
\includegraphics[width=\linewidth]{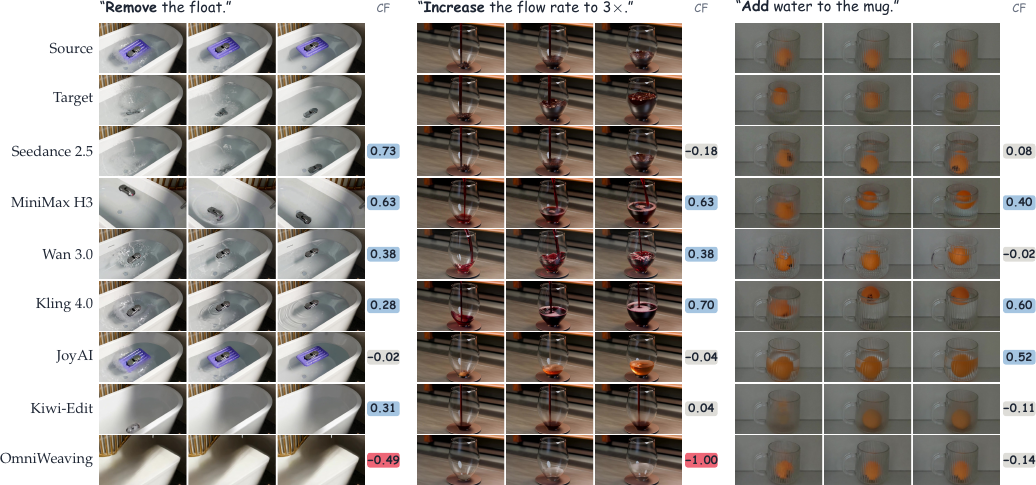}
\caption{\textbf{Qualitative comparison.} Each block shows one case with the CF of each generated video, and the third case comes from the real set. Videos that miss the edit keep the toy car afloat after removing the float, change the color of the wine instead of its flow rate, or leave the ball at the bottom of the mug after adding water.}
\label{fig:qualitative}
\end{figure}

\subsection{Main Results}
\label{sec:results}
\textbf{Physical metrics.}
Table~\ref{tab:main} shows that the judge rates the videos of all seven models as physically plausible, with a PC above 4, but finds that they rarely follow the edit, with an EA of at most 2.20.
On CF, even the best model reproduces only about one seventh of the physical change from the source video to the target video.
The commercial models lead on these metrics, with MiniMax H3, Seedance 2.5, and Wan 3.0 in the first three places on CF and EA, while JoyAI leads the open-source models on CF, and Kiwi-Edit and OmniWeaving obtain a negative CF.
CF and EA order the models in the same way in Figure~\ref{fig:radar}, although CF comes from measured quantities and EA from the ratings of the judge.
The two scores also correlate across videos, since the mean CF rises with every step of EA in Figure~\ref{fig:analysis}a.

\textbf{Visual fidelity.}
Seedance 2.5 leads on all four visual fidelity metrics, and the open-source Kiwi-Edit ranks second on PSNR, SSIM, and LPIPS.
Mask IoU, PSNR, SSIM, and LPIPS, however, favor the unedited source video over all seven models, because it matches the target video wherever the edit has no effect.
These metrics also rank the models differently from CF and EA, as MiniMax H3 reaches the outer ring of Figure~\ref{fig:radar} on CF and EA but the inner ring on SSIM.
Evaluating physical edits therefore requires measuring the physical change itself rather than the similarity to the target video.

\textbf{Agreement with human preference.}
MiniMax H3, Seedance 2.5, and Wan 3.0 also lead on HPWR in Figure~\ref{fig:radar}, whereas OmniWeaving comes last.
Figure~\ref{fig:agreement} shows that CF and EA order the models as HPWR does except for the first two, whose HPWR differs by only 0.4 points, and both metrics reach a Kendall $\tau$ of 0.90.
Mask IoU, PSNR, SSIM, and LPIPS rank the models differently, with a Kendall $\tau$ of at most 0.43.

\subsection{Analysis}
\label{sec:analysis}
\textbf{Edit types.}
Removals are the edits that the models follow best, because a removal shows in a single frame, but even here they seldom complete the edit.
Six of the seven models reach an EA of only about 3, which marks a change with one substantial mismatch, and all seven models, as well as VOID, a model built for removal, reproduce less than two fifths of the change on CF.
In the simulated videos that remove the object, six of the seven models and VOID reach a mean CF of at least 0.89 on the removed object but only $-0.08$ to 0.09 on the other objects.
The models mostly remove only the appearance of the object, so the other objects keep moving as in the source video although the removed object no longer supports, blocks, or hits them, as Figure~\ref{fig:qualitative} shows.

Insertions challenge the models more than removals, and no model exceeds a CF of 0.173 or an EA of 2.43 on them.
An insertion is already harder than a removal as a visual edit, because a model must create a new object whose appearance, position, and lighting fit the scene in every frame, whereas a removal fills in the background behind an existing object.
The model must then also infer how the new object moves and acts on the other objects, and this effect often goes missing, as when the water added in Figure~\ref{fig:qualitative} leaves the ball at the bottom of the mug.

\begin{figure}[!htbp]
\centering
\includegraphics[width=0.58\linewidth]{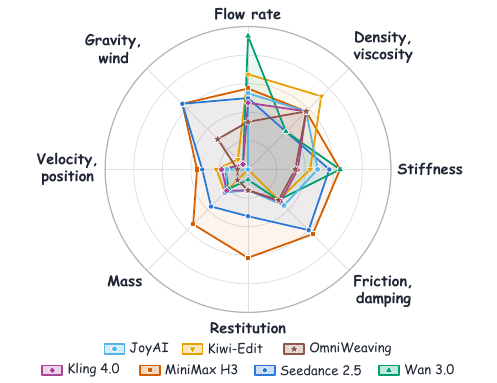}
\caption{\textbf{EA of the seven models on changes of each physical quantity.} All axes share one scale, from an EA of 1 at the center to 2.25 on the outer ring.}
\label{fig:quantity}
\end{figure}

Changes of a physical quantity are the edits that the models follow least.
No model exceeds a CF of 0.019 or an EA of 1.69 on changes, and Figure~\ref{fig:analysis}b shows that 79\% of the videos generated for a change score between $-0.25$ and 0.25, close to the behavior of the source video.
A change of a physical quantity usually leaves the scene at the execution frame as it is, so a model must infer the resulting motion from what the quantity means.

\textbf{Physical quantities.}
In Figure~\ref{fig:quantity}, every model has a higher mean EA over the flow rate, the density or viscosity, and the stiffness, which show in the shape of a liquid or an elastic object, than over the mass and the velocity or position, which show only in the trajectory of an object.
Yet CF on changes stays at most 0.019, so the models mostly change how a material looks, as with the color of the wine in Figure~\ref{fig:qualitative}, rather than how it moves.

\textbf{Interaction patterns.}
The ranking of the models by CF changes with the materials in a scene.
In Table~\ref{tab:breakdown}, Wan 3.0 leads on the rigid-rigid cases but on none of the four patterns with an elastic object or a fluid.
How well a model follows physical edits thus depends on the materials, so rigid-body scenes alone cannot rank the models.

\textbf{Execution frame at $t$.}
When a removal takes effect at a later execution frame $t$, the models often remove the object from the first frame.
In 44 of the 63 videos generated for the nine removals at $t$, the object is already gone before $t$, and every model removes it early in at least four of the nine cases.
Current models thus still struggle to apply an edit at a given execution frame $t$.

\begin{table}[!htbp]
\centering
\caption{\textbf{Results on classic rigid-body motion.} Bold and underline mark the best and second best.}
\label{tab:videophysedit}
\small
\setlength{\tabcolsep}{5pt}
\begin{tabularx}{0.78\linewidth}{l>{\centering\arraybackslash\hsize=0.94\hsize}X>{\centering\arraybackslash\hsize=1.24\hsize}X>{\centering\arraybackslash\hsize=0.94\hsize}X>{\centering\arraybackslash\hsize=0.94\hsize}X>{\centering\arraybackslash\hsize=0.94\hsize}X}
\toprule
\multirow{2}{*}[-0.55ex]{\textbf{Model}} & \multicolumn{2}{c}{\textbf{Physical metrics}} & \multicolumn{3}{c}{\textbf{Visual fidelity}} \\
\cmidrule(lr){2-3}\cmidrule(lr){4-6}
 & \textbf{CF}\,$\uparrow$ & \mbox{\textbf{Mask IoU}\,$\uparrow$} & \textbf{PSNR}\,$\uparrow$ & \textbf{SSIM}\,$\uparrow$ & \textbf{LPIPS}\,$\downarrow$ \\
\midrule
JoyAI & 0.076 & 0.297 & 24.13 & 0.761 & 0.233 \\
Kiwi-Edit & 0.047 & 0.285 & 24.69 & 0.799 & 0.135 \\
OmniWeaving & $-$0.376 & 0.158 & 23.28 & 0.789 & 0.290 \\
Kling 4.0 & 0.099 & \underline{0.366} & 22.67 & 0.766 & 0.143 \\
MiniMax H3 & \underline{0.145} & 0.191 & 22.23 & 0.750 & 0.191 \\
Seedance 2.5 & 0.060 & 0.137 & \textbf{27.46} & \textbf{0.890} & \textbf{0.108} \\
Wan 3.0 & 0.108 & 0.262 & 25.97 & \underline{0.848} & \underline{0.122} \\
VideoPhysEdit & \textbf{0.481} & \textbf{0.421} & \underline{26.15} & 0.842 & 0.161 \\
\bottomrule
\end{tabularx}
\end{table}

\textbf{Physics-based editing.}
VideoPhysEdit~\cite{Yue2026VideoPhysEdit} edits a video by simulating the edit in a scene reconstructed from the source video.
On the 10 simulated cases with classic rigid-body motion in Table~\ref{tab:videophysedit}, it reaches a CF of 0.481, whereas none of the seven models exceeds 0.145.
Its advantage comes from the eight changes of a physical quantity, on which it reaches a CF of 0.432, whereas none of the seven models exceeds 0.088.
Explicit simulation thus follows physical edits well, although it currently covers only specific materials and motions.

\section{Conclusion}
\label{sec:conclusion}
We introduced PhyCoVE, a benchmark for physical counterfactual video editing whose 533 cases pair source videos and physical edits with counterfactual target videos across rigid, elastic, and fluid dynamics.
Counterfactual fidelity scores a generated video by how much of the change toward the target video it reproduces and how well it keeps the source video before the execution frame, and it ranks the models nearly as human preference does.
Even the best of seven current video editing models reproduces only about one seventh of the physical change that an edit causes.
The models change how a scene looks but rarely how it evolves, whereas a physical edit requires understanding the dynamics that it changes.
They also often ignore when the edit takes effect.
Explicit simulation computes these dynamics but covers only specific materials and motions.
Physical counterfactual video editing thus remains far from solved, and PhyCoVE provides a foundation for developing and evaluating methods for it.

\FloatBarrier
\bibliographystyle{plainnat}
\bibliography{references}

\clearpage
\appendix
\setcounter{equation}{0}
\setcounter{figure}{0}
\setcounter{table}{0}
\renewcommand{\theequation}{S\arabic{equation}}
\renewcommand{\thefigure}{S\arabic{figure}}
\renewcommand{\thetable}{S\arabic{table}}
\renewcommand{\theHequation}{S\arabic{equation}}
\renewcommand{\theHfigure}{S\arabic{figure}}
\renewcommand{\theHtable}{S\arabic{table}}
\addtocontents{toc}{\protect\setcounter{tocdepth}{2}}
\begingroup
\renewcommand{\contentsname}{Table of Contents for Appendix}
\setstretch{1}
\fontsize{10.5}{14}\selectfont
\renewcommand{\cfttoctitlefont}{\fontsize{16}{20}\selectfont\bfseries\color{seedblue}}
\renewcommand{\cftsecfont}{\fontsize{11.5}{14.5}\selectfont\bfseries}
\renewcommand{\cftsecpagefont}{\fontsize{11.5}{14.5}\selectfont\bfseries}
\renewcommand{\cftsubsecfont}{\fontsize{10.5}{13.5}\selectfont}
\renewcommand{\cftsubsecpagefont}{\fontsize{10.5}{13.5}\selectfont}
\setlength{\cftbeforetoctitleskip}{0pt}
\setlength{\cftaftertoctitleskip}{12pt}
\setlength{\cftbeforesecskip}{7pt}
\setlength{\cftbeforesubsecskip}{2pt}
\setlength{\cftparskip}{0pt}
\cftsetindents{section}{0em}{1.8em}
\cftsetindents{subsection}{1.8em}{2.7em}
\cftsetpnumwidth{2em}
\cftsetrmarg{3em}
\tableofcontents
\endgroup
\clearpage
\raggedbottom
\section{Counterfactual Fidelity}
\label{sec:supp_cf}
This section gives the full protocol of counterfactual fidelity (CF), which Section~\ref{sec:metrics} of the main paper defines by Eq.~\eqref{eq:cf}.

\subsection{Videos and Objects}
\label{sec:supp_cf_objects}
We compare every generated video with the source and target videos of its case on the 96 frames at 24 fps of the case.
We resample every generated video to this timeline by its timestamps.
We locate every object of a simulated case from the 3D positions that the simulator records, projected into the image, and every object of a real case from points and boxes that humans place on the source and target videos.
We follow each object with CoTracker3~\cite{Karaev2024CoTracker3}, seeded with points on the object, and with SAM~2~\cite{Ravi2025SAM2}, prompted on the object, and measure positions at the $1280\times720$ resolution.
For each object of a simulated case, we use the tracker that stays within $\max(16, 0.6r)$ pixels of the projected position in most frames of the source and target videos, where $r$ is the radius of the object, and when neither tracker holds the object, we prompt SAM~2 again at its projected position.
We segment a liquid as the region in which a video differs from its own first frame in CIELAB color, within the area that the liquid reaches in the source or target video and after matching the tone of the background.

\subsection{Measured Quantities}
\label{sec:supp_cf_quantities}
From the track of an object we take its displacement from its own first frame and measure six quantities, namely the final displacement, the length of the path, the horizontal range, the lowest and the highest point, and the time at which the object comes to rest.
For an elastic object we add its height, width, and area at the end of the clip and its smallest height, all measured on its SAM~2 mask.
For a liquid we measure its area at one quarter, one half, three quarters, and the end of the clip, the number of frames that show a stream, and the top, bottom, and width of the liquid at the end.
For a removed object we measure whether it is gone after the execution frame, and for an inserted object whether it is present at its position in the target video.

\subsection{Scoring}
\label{sec:supp_cf_scoring}
We score a quantity only when the target video changes it by more than a minimum, which is 8 to 16 pixels for positions and sizes, 4 to 6 frames for the time of rest and the stream, and at least a tenth of the value for lengths and areas.
Except for the time of rest, every quantity of the track must also change by more than three times the tracking noise of the object, which we measure between the target video and its re-encoded copy.
The change of a quantity must also be measurable, that is, on re-encoded copies of the source and target videos the measurement must recover it, with a score of at least 0.5 for the copy of the target video and within $\pm0.5$ for the copy of the source video.
We score each quantity by Eq.~\eqref{eq:cf}, where a vector quantity such as the final displacement enters through its inner product, a score above 1 loses its excess over 1 and becomes $2-\mathrm{CF}(q)$, so it falls to 0 at twice the change, and we clip all scores to $[-1,1]$.
We score the presence of a removed or inserted object by where the generated video falls between the source and target videos on blurred CIELAB crops at the position of the object, from 0 when it looks like the source video to 1 when it looks like the target video.
Every quantity of an object scores $-1$ when the generated video drops the object while the target video still shows it, or keeps the object still although it moves in both the source and target videos, and a video that does not perform a requested removal or insertion scores at most 0 on every quantity of that object.
The CF of a video is the mean over its scored quantities, and the CF of a model is the mean over its scored videos of both the simulated and the real cases.
CF covers the 506 cases in which the edit changes at least one quantity measurably.

\subsection{Verification and Reliability}
\label{sec:supp_cf_verify}
Generated videos contain fast motion, occlusion, and objects that change their appearance, duplicate, or vanish, so humans check every track against its video on sheets that show the object and its track across the clip.
They correct a track that drifts to another object or loses its target by prompting SAM~2 again on the object, follow an object again where it reappears, and mark an object that the video does not show as absent.
They leave out an object that the video still shows but that cannot be measured, and a video in which no scored object remains does not enter CF.
In the real cases, humans also judge the presence of removed and inserted objects and the strength of a water stream directly on the videos.
Each such judgment scores 1 when the generated video matches the target video and 0 when it matches the source video.
Under this protocol, 94\% of the quantities that the target videos of the scored simulated cases change beyond the minimum pass the check on the re-encoded copies, and on the scored quantities the re-encoded copy of the target video obtains a CF of 0.960 and that of the source video 0.004.

\subsection{Edits at a Later Frame}
\label{sec:supp_cf_later}
The target video of a case equals its source video before the execution frame $t$, so a generated video that changes the scene before this frame departs from the target video, and we deduct this departure.
For every scored object, we measure its departure from the source video in each frame, $d(f)$, as the distance between the displacements of the object since its first frame for a tracked object, as its absence for a removed object, and as the difference in area for a liquid.
Before the execution frame we count only departures beyond the tolerance $\max(16, 0.6r)$ of Section~\ref{sec:supp_cf_objects} for a tracked object and beyond $\max(480, 0.1a)$ square pixels for a liquid, where $a$ is the largest area of the liquid in the source video.
With $E=\sum_{f<t} d^{\mathrm{gen}}(f)$ for the generated video and $G=\sum_{f\geq t} d^{\mathrm{tgt}}(f)$ for the target video, each score $\mathrm{CF}(q)$ of the object in Eq.~\eqref{eq:cf} of the main paper becomes $\mathrm{CF}(q)-E/G$, clipped to $[-1,1]$.
A video that keeps the source video until the execution frame loses nothing, so the rule leaves every edit at the first frame and every correctly timed video unchanged.
The deduction grows with how early and how strongly a video changes the scene, so removing an object two frames early costs little, whereas removing it from the first frame cancels the credit for the removal when the edit falls in the middle of the clip.
Both sums use the same per-frame departure, so the ratio has no unit and applies alike to rigid objects, elastic objects, and liquids.
Humans verify every departure of a removed object or a liquid on the frames before the execution frame, as they verify every track.

\subsection{Mask IoU and Visual Fidelity}
\label{sec:supp_cf_other}
For Mask IoU~\cite{Yue2026VideoPhysEdit}, we follow every solid object whose motion enters CF, segment it in the generated and target videos along the same verified tracks, and compute the intersection over union of the two regions in every frame in which the target video shows the object, so an object that the generated video drops scores 0.
The Mask IoU of a video is the mean over its frames and objects, and liquids enter only CF.
PSNR, SSIM~\cite{Wang2004SSIM}, and LPIPS~\cite{Zhang2018LPIPS} compare each generated video with its target video at $1280\times720$ on the same 24 fps timeline as CF.
FVD~\cite{Unterthiner2018FVD} compares the Kinetics-400 I3D features of the videos of a model with those of the target videos of the same cases.

\section{Judge Scores}
\label{sec:supp_judge}
This section gives the protocol of the two judge scores, edit adherence (EA) and physical commonsense (PC), that Section~\ref{sec:metrics} of the main paper defines.

\subsection{Inputs}
\label{sec:supp_judge_inputs}
The judge is GPT-6 Astra~\cite{OpenAI2026GPT6Astra} at medium reasoning effort, and we decode every frame of a video and lay the frames out in time order on sheets.
Each frame on a sheet carries its frame number and its time in seconds, and the prompt asks the judge to compare videos at the same displayed time.
For EA, the judge receives the instruction and every frame of the source, target, and generated videos, with the role of each video written on its frames.
For PC, the judge receives only the generated video.
For an edit at a later frame, the EA prompt states the time of the execution frame, and the PC sheets start at that frame, which the judge treats as the initial state.
The judge uses no tools and returns an integer score from 1 to 5 with a short justification that cites visible objects and times.
The released evaluation code contains the full prompts.
We score every video that the seven models and VOID~\cite{Motamed2026VOID} return.

\subsection{Scoring Rules}
\label{sec:supp_judge_rules}
For EA, the judge first identifies the change from the source video to the target video and then checks whether the generated video shows this change on the same object, with the same direction, onset, magnitude, and direct consequences.
It accepts physically equivalent trajectories and splashes, judges a change of a physical quantity by its visible outcome, and counts a change that comes early, late, or only briefly as a timing error.
When the judge finds no visible change from the source video to the target video, the expected response is no visible change, and EA measures how well the generated video keeps the visible state and motion of the source video.
For PC, the judge rates motion continuity, contact, deformation, flow, static support, and the identity of each object over time.
Table~\ref{tab:supp_judge_scales} gives the scales.
Each score of a model averages over its videos with equal weight, and Joint is the percentage of videos with an EA and a PC of at least 4.
Section~\ref{sec:supp_human_agreement} compares EA with human preference.

\begin{table}[!htbp]
\centering
\caption{\textbf{Scales of the judge scores.} EA follows the first scale when the judge finds a visible change from the source video to the target video and the second scale otherwise.}
\label{tab:supp_judge_scales}
\small
\setlength{\tabcolsep}{5pt}
\renewcommand{\arraystretch}{1.15}
\begin{tabularx}{\linewidth}{c>{\raggedright\arraybackslash}X>{\raggedright\arraybackslash}X>{\raggedright\arraybackslash}X}
\toprule
\textbf{Score} & \textbf{EA, target change visible} & \textbf{EA, target change not visible} & \textbf{PC} \\
\midrule
5 & All requested changes and their consequences match the target video. & The generated video keeps the visible state, motion, and consequences of the edited object. & No clear physical anomaly is visible. \\
4 & A minor deviation in timing or magnitude. & Only minor deviations. & A small local anomaly. \\
3 & The right object and direction but one substantial mismatch. & One substantial visible mismatch. & A clear anomaly, but the main process is plausible. \\
2 & A weak or transient response. & Major mismatches with some matching aspects. & A major contradiction in a key interaction. \\
1 & No change where a clear change is required, a wrong object, or a reversed change. & The main response contradicts the target video, for example by removing or replacing the object or changing its motion outcome. & The principal event is physically incoherent. \\
\bottomrule
\end{tabularx}
\end{table}

\section{Human Preference Study}
\label{sec:supp_human}
This section gives the protocol and the full results of the human preference study that Section~\ref{sec:metrics} of the main paper summarizes by the human preference win rate (HPWR).

\subsection{Protocol}
\label{sec:supp_human_protocol}
Each comparison shows the instruction of a case, its source video, its target video, and two generated videos, all played in sync at their original frame rate.
The interface hides the model names and places the two generated videos in a random order, and an annotator watches all four videos before choosing the first video, the second video, or a tie.
The question asks which of the two videos better completes the physical edit, considering whether it follows the instruction, whether it stays consistent with the source and target videos, whether it is physically plausible, and its overall video quality.

Annotators weigh these aspects in the following order of priority.
\begin{enumerate}
\item Edit adherence. The edited object, the physical quantity, the direction, the timing, and the direct outcome follow the instruction and the target video.
\item Physical plausibility. Rigid bodies, elastic objects, fluids, their interactions, and the identity of each object evolve consistently.
\item Preservation and quality. Content unrelated to the edit stays stable, with no vanishing objects, texture changes, flicker, or jumps in time.
\end{enumerate}
For an edit at a later frame, the interface marks the execution frame.
Annotators choose a tie when they cannot reliably tell the two videos apart.
Ten annotators take part, and three different annotators judge every comparison independently.

\subsection{Coverage}
\label{sec:supp_human_coverage}
The study covers a balanced panel of 500 comparisons on 71 of the 527 cases for which all seven models return a video, so that every pair of models appears 23 or 24 times.
We draw the 71 cases by a seeded random selection under fixed quotas for each edit type, interaction pattern, set, and execution frame, with at most one case per scene and edit type.
Each of the ten annotators judges 150 comparisons, which gives 1{,}500 judgments.
VOID~\cite{Motamed2026VOID} returns videos for removals only and does not enter the study.
Table~\ref{tab:supp_human_coverage} lists the cases, comparisons, and judgments of each edit type.
Insertions make up 15 of the 71 cases but only 21 of the 533 cases of PhyCoVE, so the overall results weight each judgment by the number of cases of its edit type in PhyCoVE divided by that in the study, with 178 removals, 334 changes, and 21 insertions in PhyCoVE.

\begin{table}[!htbp]
\centering
\caption{\textbf{Coverage of the human preference study.} A comparison pits the videos of two models on one case and receives three judgments.}
\label{tab:supp_human_coverage}
\small
\setlength{\tabcolsep}{8pt}
\begin{tabular*}{0.62\linewidth}{@{\extracolsep{\fill}}lccc}
\toprule
\textbf{Edit type} & \textbf{Cases} & \textbf{Comparisons} & \textbf{Judgments} \\
\midrule
Remove & 24 & 168 & 504 \\
Change & 32 & 227 & 681 \\
Insert & 15 & 105 & 315 \\
\midrule
Total & 71 & 500 & 1{,}500 \\
\bottomrule
\end{tabular*}
\end{table}

\subsection{Results}
\label{sec:supp_human_results}
We report two scores for each model in four subsets, namely all comparisons and the comparisons of each edit type.
Within each edit type, HPWR is the share of points that a model earns over the judgments of its comparisons, with one point for a win, half a point for a tie, and none for a loss, so an HPWR of 50\% marks parity with the other models.
For the Bradley--Terry score with ties (BTT), we fit the Rao--Kupper extension~\cite{Rao1967Ties} of the Bradley--Terry model~\cite{Bradley1952RankAnalysis} to the wins, ties, and losses and scale the strongest model of each subset to 100.
The overall scores combine the edit types with the weights of Section~\ref{sec:supp_human_coverage}.

\begin{figure}[!htbp]
\centering
\includegraphics[width=\linewidth]{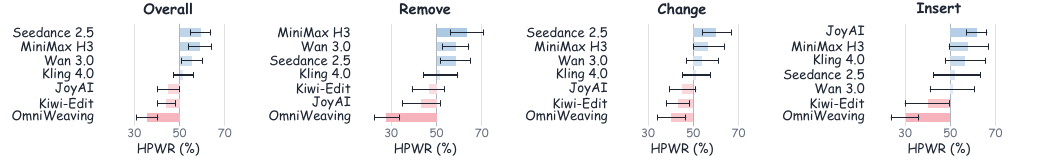}
\caption{\textbf{HPWR of every model in each subset.} The models run from the first to the last rank, the bars start at an HPWR of 50\%, and the lines give the 95\% bootstrap confidence intervals~\cite{Efron1993Bootstrap}.}
\label{fig:supp_human_bars}
\end{figure}

Table~\ref{tab:supp_human_results} and Figure~\ref{fig:supp_human_bars} give the scores of every model.
HPWR and BTT give the same order in every subset.
Seedance 2.5 ranks first overall and on changes, MiniMax H3 on removals, and JoyAI on insertions.

\begin{table}[!htbp]
\centering
\caption{\textbf{Human preference scores.} HPWR is in percent, and bold marks the first rank of each subset.}
\label{tab:supp_human_results}
\small
\setlength{\tabcolsep}{4pt}
\begin{tabularx}{\linewidth}{l*{8}{>{\centering\arraybackslash}X}}
\toprule
\multirow{2}{*}[-0.55ex]{\textbf{Model}} & \multicolumn{2}{c}{\textbf{Overall}} & \multicolumn{2}{c}{\textbf{Remove}} & \multicolumn{2}{c}{\textbf{Change}} & \multicolumn{2}{c}{\textbf{Insert}} \\
\cmidrule(lr){2-3}\cmidrule(lr){4-5}\cmidrule(lr){6-7}\cmidrule(lr){8-9}
 & HPWR & BTT & HPWR & BTT & HPWR & BTT & HPWR & BTT \\
\midrule
JoyAI & 44.9 & 59.4 & 43.1 & 47.3 & 44.9 & 57.8 & \textbf{61.7} & \textbf{100.0} \\
Kiwi-Edit & 44.1 & 57.9 & 46.5 & 53.6 & 43.1 & 54.4 & 40.0 & 45.2 \\
OmniWeaving & 35.8 & 42.4 & 27.8 & 25.9 & 40.3 & 48.9 & 30.6 & 31.4 \\
Kling 4.0 & 51.6 & 75.8 & 51.7 & 64.9 & 51.3 & 72.9 & 56.7 & 82.5 \\
MiniMax H3 & 59.0 & 98.5 & \textbf{63.5} & \textbf{100.0} & 56.7 & 88.2 & 57.8 & 85.9 \\
Seedance 2.5 & \textbf{59.4} & \textbf{100.0} & 58.7 & 83.6 & \textbf{60.2} & \textbf{100.0} & 52.2 & 70.3 \\
Wan 3.0 & 55.3 & 86.1 & 58.7 & 83.6 & 53.8 & 79.2 & 51.1 & 67.6 \\
\bottomrule
\end{tabularx}
\end{table}

\subsection{Agreement with Automatic Metrics}
\label{sec:supp_human_agreement}
We compare the order that each metric of Table~\ref{tab:main} gives the seven models with the order of HPWR, using the Kendall $\tau$~\cite{Kendall1938Tau} in its $\tau_b$ form for ties~\cite{Kendall1945Ties} and the Spearman $\rho$~\cite{Spearman1904Rho}, both on the unrounded values, with the signs of LPIPS and FVD reversed so that a higher value is better for every metric.
For the edit types, we compare CF and EA on each edit type with HPWR on the same edit type.
With seven models, swapping one pair of adjacent models lowers $\tau$ from 1 to 0.90.

\begin{figure}[!htbp]
\centering
\includegraphics[width=\linewidth]{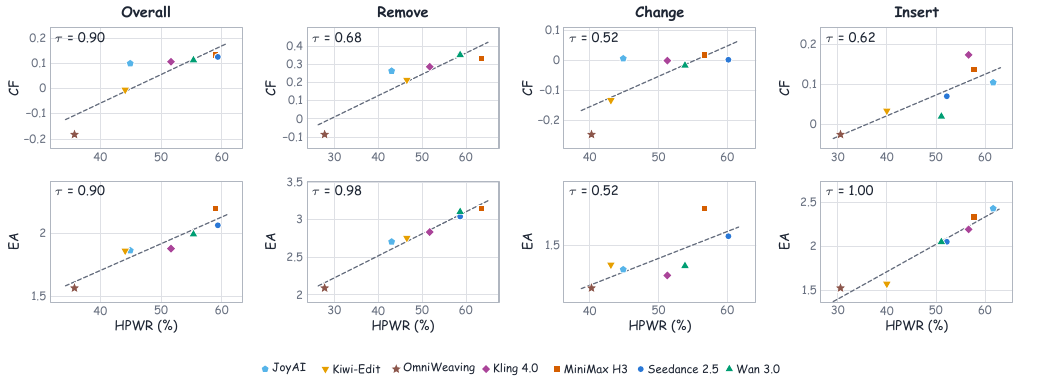}
\caption{\textbf{CF and EA against HPWR in each subset.} Each panel gives the Kendall $\tau$ over the seven models, and the dashed line is a least-squares fit over them.}
\label{fig:supp_human_scatter}
\end{figure}

CF and EA order the seven models as HPWR does except for Seedance 2.5 and MiniMax H3, whose HPWR differs by only 0.4 points and whose 95\% intervals in Figure~\ref{fig:supp_human_bars} almost coincide.
Joint matches the order of HPWR exactly, whereas Mask IoU, PSNR, SSIM, and LPIPS reach a $\tau$ of at most 0.43.
Within each edit type, CF and EA keep a positive $\tau$ with HPWR, as Figure~\ref{fig:supp_human_scatter} shows.

\section{Benchmark Details}
\label{sec:supp_benchmark}
This section details the construction, data format, and statistics of PhyCoVE beyond Section~\ref{sec:benchmark} of the main paper.

\subsection{Construction of the Simulated Set}
\label{sec:supp_benchmark_sim}
\textbf{Scene programs.}
The program that the coding agents write for a scene builds it from 3D assets and also simulates and renders every edit of the scene.
We take the assets from Poly Haven, ambientCG, BlenderKit, and Sketchfab or make them for the scene.

\textbf{Solvers.}
Each program simulates every material with a solver suited to it.
Most rigid-rigid programs simulate their objects with PyBullet~\cite{Coumans2016PyBullet}, and the others with MuJoCo~\cite{Todorov2012MuJoCo} or a custom rigid-body solver.
Elastic objects use incremental potential contact (IPC)~\cite{Li2020IPC}, the material point method (MPM)~\cite{Hu2018MLSMPM}, the finite element solver of Genesis~\cite{Genesis2024Genesis}, or a custom elastic model.
Liquids mostly use the smoothed-particle hydrodynamics (SPH) solver~\cite{Monaghan1992SPH} of Genesis or the fluid-implicit-particle (FLIP) solver~\cite{Zhu2005Sand} of Blender~\cite{Blender2024}, and Genesis couples its SPH liquids with its MPM solids in both directions, so an elastic object deforms under a liquid and displaces it in turn.
The quantities that the edits change are inputs of these simulations, such as the mass, friction, restitution, or initial velocity of a rigid object, the stiffness of an elastic object, or the density, viscosity, or flow rate of a liquid.

\textbf{Rendering.}
We render every scene with physically based rendering in Blender~\cite{Blender2024} at $1280\times720$ pixels and 24 fps for 96 frames and encode the videos with H.264.
For a liquid simulated with SPH, we reconstruct its surface from the particles at every frame and render this surface as a mesh.

\textbf{Counterfactual simulation.}
A change sets an input of the simulation to its new value, a removal deletes the object or fluid from the simulation and the scene from the execution frame on, and an insertion adds a new object to the simulation on the line between two objects or points of the scene, at its midpoint or at a given fraction of its length.
For an edit at the first frame, the program applies the edit to the initial scene and simulates the whole clip.
For an edit at a later frame, the program either changes the input at the execution frame within the simulation of the source scene or restores the source scene state saved at that frame, applies the edit, and continues the simulation.
We compare the exported states of all 63 cases with a later execution frame against those of their source scenes and find them identical frame by frame before the execution frame.

\subsection{Construction of the Real Set}
\label{sec:supp_benchmark_real}
\textbf{Recording.}
We record each source group of the real set continuously with a phone at one position, first the source scene and then the scene with each edit, which we set up by hand by removing one or two objects or the water, inserting an object or water, or turning up the flow of a faucet.
We write the instruction of each real case by hand and name every object in it by its appearance.

\textbf{Trimming and alignment.}
From the stretches of each recording in which no person or hand appears in the frame, we cut one take per video, so all takes of a group share the camera position and the environment.
We bring all takes of a group to $1280\times720$ pixels in the same way.
Every output frame at 24 fps takes the recorded frame nearest to its timestamp.
We anchor the takes of a group at an event that they share, mostly the frame at which the moving object enters the view, so this event falls on the same frame in every take.

\textbf{Exposure.}
The automatic exposure of the phone shifts the brightness of a take when an object enters the view, so the takes of a group can differ in brightness at the same frame.
We remove these shifts in the eight of the 45 takes that show them.
For each frame and color channel of such a take, we fit a gain and a gamma that map the static background of the frame onto the per-pixel temporal median of the take, smooth both over three frames, and apply them to the whole frame.

\textbf{Annotations.}
Humans mark the frames of the key events in each take, such as the entry of the moving object and its first contact, and place points and boxes on every object in the source and target videos.

\subsection{Data Format and Release}
\label{sec:supp_benchmark_format}
\textbf{Layout.}
The benchmark stores each source group in a folder \texttt{scenes/}\textit{scene}\texttt{/cases/}, in which the subfolder \texttt{source} holds the source video and every other subfolder holds one case.
Each subfolder contains its video in \texttt{video.mp4} and its states or annotations in \texttt{ground\_truth\_transforms.json}, and the subfolder of a case also gives its instruction and edit specification in \texttt{prompts.json}.
The simulated set occupies the root of the benchmark, and the real set occupies the folder \texttt{real/} with the same layout.

\textbf{Metadata.}
A metadata file lists every case with its source group, interaction pattern, edit specification, edited object, execution frame, and instruction.

\textbf{Edit specification.}
The edit specification states each edit in a form that programs can read.
\texttt{SET} $o.q$ \texttt{TIMES} $k$ scales the quantity $q$ of the object or scene $o$ by the factor $k$, and \texttt{SET} $o.q$ \texttt{TO} $v$ sets it to the value $v$.
\texttt{DELETE} $o$ removes $o$, and \texttt{ADD} $o$ \texttt{BETWEEN} $a$ \texttt{AND} $b$ inserts $o$ on the line between the objects or points $a$ and $b$, at the position that \texttt{AT MIDPOINT} or a fraction such as \texttt{AT 1/3 FROM} $b$ gives.
A trailing \texttt{AT FRAME} $t$ sets a later execution frame $t$, and a template turns every specification of the simulated set into its instruction.
The real set adds three forms, which remove two objects at once, insert an object by name or at a named region, and increase a flow without a factor, as in \texttt{SET water.flow\_rate INCREASE}.

\textbf{States and annotations.}
For the simulated set, \texttt{ground\_truth\_transforms.json} gives the intrinsics of the camera and its pose at each of the 96 frames.
At each frame, it also gives for every rigid and elastic object whether the object is present, its position in meters, its orientation as a quaternion, and its transformation matrix.
These values give the pose of a rigid object and follow a representative point of an elastic object, such as the centroid of its surface, and with the camera parameters every position projects into the image.
For the real set, the same file gives the recorded frame that each frame takes, whether each object is present in the take, and the annotated events.

\textbf{Release.}
We will release the benchmark in this format, with its 110 source videos, 533 target videos, and the instructions, edit specifications, metadata, states, and annotations described above, together with the evaluation code.

\subsection{Statistics}
\label{sec:supp_benchmark_stats}
Table~\ref{tab:supp_benchmark_cases} shows that changes of a physical quantity make up about two thirds of the simulated set and that both the changes and the edits at a later frame span all five interaction patterns.
Each simulated source group contains 3 to 10 cases with a median of 6, and each real source group contains 1 or 2 cases.
The 63 edits at a later frame comprise 54 changes and 9 removals, and their execution frames range from frame 3 to frame 73.
Table~\ref{tab:supp_benchmark_quantities} shows that every interaction pattern contains changes of several quantity groups and that the groups follow the materials, with restitution only in the rigid-rigid scenes, stiffness only in the scenes with an elastic object, and the flow rate only in the scenes with a liquid.

\begin{table}[!htbp]
\centering
\caption{\textbf{Cases by interaction pattern.} R, E, and F mean rigid, elastic, and fluid, Groups counts the source groups, and First and Later count the edits whose execution frame is the first frame or a later frame.}
\label{tab:supp_benchmark_cases}
\small
\setlength{\tabcolsep}{5pt}
\begin{tabular*}{0.88\linewidth}{@{\extracolsep{\fill}}lccccccc}
\toprule
\multirow{2}{*}[-0.55ex]{\textbf{Pattern}} & \multirow{2}{*}[-0.55ex]{\textbf{Groups}} & \multicolumn{3}{c}{\textbf{Edit type}} & \multicolumn{2}{c}{\textbf{Execution frame}} & \multirow{2}{*}[-0.55ex]{\textbf{Cases}} \\
\cmidrule(lr){3-5}\cmidrule(lr){6-7}
 & & Remove & Change & Insert & First & Later & \\
\midrule
\multicolumn{8}{l}{\textit{Simulated set}} \\
R-R & 40 & 69 & 162 & 5 & 201 & 35 & 236 \\
R-E & 28 & 55 & 93 & 12 & 150 & 10 & 160 \\
R-F & 16 & 25 & 55 & 0 & 66 & 14 & 80 \\
E-F & 4 & 5 & 16 & 0 & 18 & 3 & 21 \\
E-E & 2 & 3 & 7 & 1 & 10 & 1 & 11 \\
Total & 90 & 157 & 333 & 18 & 445 & 63 & 508 \\
\midrule
\multicolumn{8}{l}{\textit{Real set}} \\
R-R & 10 & 12 & 0 & 1 & 13 & 0 & 13 \\
R-E & 4 & 4 & 0 & 0 & 4 & 0 & 4 \\
R-F & 4 & 3 & 1 & 1 & 5 & 0 & 5 \\
E-E & 2 & 2 & 0 & 1 & 3 & 0 & 3 \\
Total & 20 & 21 & 1 & 3 & 25 & 0 & 25 \\
\midrule
\multicolumn{8}{l}{\textit{PhyCoVE}} \\
R-R & 50 & 81 & 162 & 6 & 214 & 35 & 249 \\
R-E & 32 & 59 & 93 & 12 & 154 & 10 & 164 \\
R-F & 20 & 28 & 56 & 1 & 71 & 14 & 85 \\
E-F & 4 & 5 & 16 & 0 & 18 & 3 & 21 \\
E-E & 4 & 5 & 7 & 2 & 13 & 1 & 14 \\
Total & 110 & 178 & 334 & 21 & 470 & 63 & 533 \\
\bottomrule
\end{tabular*}
\end{table}
\FloatBarrier

\begin{table}[!htbp]
\centering
\caption{\textbf{Changes of a physical quantity by quantity group.} The groups follow Figure~\ref{fig:quantity} of the main paper, the counts cover both sets, and Later counts the changes at a later frame.}
\label{tab:supp_benchmark_quantities}
\small
\setlength{\tabcolsep}{5pt}
\begin{tabular*}{0.90\linewidth}{@{\extracolsep{\fill}}lccccccc}
\toprule
\multirow{2}{*}[-0.55ex]{\textbf{Quantity}} & \multicolumn{5}{c}{\textbf{Interaction pattern}} & \multirow{2}{*}[-0.55ex]{\textbf{Later}} & \multirow{2}{*}[-0.55ex]{\textbf{Total}} \\
\cmidrule(lr){2-6}
 & R-R & R-E & R-F & E-F & E-E & & \\
\midrule
Flow rate & 0 & 0 & 20 & 4 & 0 & 8 & 24 \\
Density, viscosity & 0 & 3 & 24 & 3 & 2 & 3 & 32 \\
Stiffness & 0 & 40 & 0 & 4 & 2 & 1 & 46 \\
Friction, damping & 31 & 11 & 0 & 0 & 3 & 11 & 45 \\
Restitution & 22 & 0 & 0 & 0 & 0 & 4 & 22 \\
Mass & 42 & 31 & 10 & 1 & 0 & 11 & 84 \\
Velocity, position & 55 & 8 & 2 & 0 & 0 & 7 & 65 \\
Gravity, wind & 12 & 0 & 0 & 4 & 0 & 9 & 16 \\
\midrule
Total & 162 & 93 & 56 & 16 & 7 & 54 & 334 \\
\bottomrule
\end{tabular*}
\end{table}
\FloatBarrier

\section{Evaluated Models}
\label{sec:supp_models}
This section gives the settings of the models that Section~\ref{sec:metrics} of the main paper evaluates.

\subsection{Common Protocol}
\label{sec:supp_models_protocol}
Every model receives the instruction of a case, which states the edit and its execution frame, and the source video at the input size and length that it supports, which Table~\ref{tab:supp_models} lists.
Open-source models run their official code and commercial models run through the APIs of their providers, with one configuration for all cases, and we score the one video that a model returns for each case.
For Seedance 2.5, we evaluate over 527 cases due to API policy restrictions.
We compute CF, Mask IoU, and the visual fidelity metrics on the 96 frames at 24 fps of a case after resampling the returned videos by their timestamps, as Section~\ref{sec:supp_cf_objects} describes.

\begin{table}[!htbp]
\centering
\caption{\textbf{Settings of the evaluated models.} Input and Output give the videos that each model receives and returns.}
\label{tab:supp_models}
\footnotesize
\setlength{\tabcolsep}{4pt}
\begin{tabularx}{\linewidth}{lll>{\raggedright\arraybackslash}X}
\toprule
\textbf{Model} & \textbf{Input} & \textbf{Output} & \textbf{Settings} \\
\midrule
\multicolumn{4}{l}{\textit{Open-source}} \\
JoyAI-Video-Edit (JoyAI)~\cite{Xiao2026JoyAIVideoEdit} & $1248\times720$, 96 frames & $1248\times720$, 96 frames, 24\,fps & FP8 inference, 2 sampling steps, seed 42 \\
Kiwi-Edit~\cite{Lin2026KiwiEdit} & $1280\times720$, 81 frames & $1280\times720$, 81 frames, 24\,fps & Instruction-only 5B model, 50 sampling steps, seed 0 \\
OmniWeaving~\cite{Pan2026OmniWeaving} & $848\times480$, 97 frames & $848\times480$, 96 frames, 24\,fps & Editing mode, full frame resized, last frame repeated, 50 sampling steps, seed 0 \\
VOID~\cite{Motamed2026VOID} & $672\times384$, 96 frames & $672\times384$, 96 frames, 24\,fps & Removal cases only, first inference pass, 30 sampling steps, seed 42 \\
\midrule
\multicolumn{4}{l}{\textit{Commercial}} \\
Kling 4.0~\cite{Kuaishou2026Kling} & $1280\times720$, 96 frames & $1280\times720$, 97 frames, 24\,fps & API, 720p, 4\,s, adaptive aspect ratio \\
MiniMax H3~\cite{MiniMax2026H3} & $1280\times720$, 96 frames & $1344\times768$, 107 frames, 24\,fps & API, 768p, 4\,s, adaptive aspect ratio \\
Seedance 2.5~\cite{ByteDance2026Seedance} & $1280\times720$, 96 frames & $1280\times720$, 89 frames, 24\,fps & API \\
Wan 3.0~\cite{Alibaba2026Wan} & $1280\times720$, 96 frames & $1280\times720$, 120 frames, 30\,fps & API \\
\midrule
\multicolumn{4}{l}{\textit{Physics-based}} \\
VideoPhysEdit~\cite{Yue2026VideoPhysEdit} & $1280\times720$, 96 frames & $720\times480$, 96 frames, 24\,fps & 10 simulated cases of classic rigid-body motion \\
\bottomrule
\end{tabularx}
\end{table}

\subsection{Settings of Each Model}
\label{sec:supp_models_settings}
We queried the commercial APIs in November 2026.
The prompt of Kling 4.0 starts with ``Edit @Video1.'', in which @Video1 denotes the input video.

VOID~\cite{Motamed2026VOID} only performs removals, so we run it on the 178 removal cases, 157 simulated and 21 real, with Gemini 3.1 Flash-Lite~\cite{Google2026GeminiFlashLite} as its vision-language model.
Besides the instruction, its pipeline takes points on each object to remove and one sentence that describes the background.
We place the points with the same vision-language model, which reads a sheet of 16 source frames together with the instruction and returns the earliest frame on which all objects to remove are visible and a box around each object on that frame, and with SAM~2~\cite{Ravi2025SAM2}, which segments each box, so that the four extreme pixels and the center of each mask become the points of its object.
We write the background sentence of each case once from its source video.

\subsection{VideoPhysEdit on Classic Rigid-Body Motion}
\label{sec:supp_models_vpe}
We run VideoPhysEdit~\cite{Yue2026VideoPhysEdit} on 10 simulated scenes of classic rigid-body motion that it supports, in which one to three rigid objects slide, roll, bounce, or topple on a flat support, with one case per scene whose edit it supports.
Table~\ref{tab:supp_models_vpe} lists the cases, of which eight change the mass, the initial velocity, or the restitution of an object and two remove an object, all from the first frame.

\begin{table}[!htbp]
\centering
\caption{\textbf{Cases of classic rigid-body motion.} Every edit takes effect at the first frame.}
\label{tab:supp_models_vpe}
\small
\setlength{\tabcolsep}{6pt}
\begin{tabularx}{\linewidth}{>{\raggedright\arraybackslash}X>{\raggedright\arraybackslash}X}
\toprule
\textbf{Scene} & \textbf{Edit} \\
\midrule
A basketball bounces on a court & Reduce the restitution of the basketball to 0.5$\times$ \\
A basketball rebounds in a room corner & Reduce the initial velocity of the basketball to 0.3$\times$ \\
Two soccer balls collide head-on & Increase the mass of the right soccer ball to 3$\times$ \\
A toy car hits a resting toy car & Increase the initial velocity of the purple toy car to 3$\times$ \\
A toy car hits a plastic bottle & Increase the initial velocity of the toy car to 1.5$\times$ \\
A bowl slides into a resting bowl & Increase the mass of the left bowl to 5$\times$ \\
A tin slides into a resting tin & Increase the mass of the round tin to 5$\times$ \\
A curling stone slides into two stones & Increase the mass of the blue-handled stone to 3$\times$ \\
A toy ball rebounds from a toy chest into a soccer ball & Remove the soccer ball \\
A rolling can hits an upright can & Remove the upright can \\
\bottomrule
\end{tabularx}
\end{table}

On each case, VideoPhysEdit runs its seven stages, which identify and track the objects, analyze their observed motion, reconstruct the scene, reconstruct a motion prior, estimate the physical parameters and initial states by simulation, apply the edit at the execution frame and simulate the rest of the video, and generate the edited video from the simulated trajectories and an edited reference image.
As in its evaluation on PCVE-RigidBench, we give its perception stage the inputs that the benchmark protocol provides, namely the categories of the moving objects and a box on the first frame where the generic detector misses an object, and we write each instruction in the format of its templates.
The comparison thus reflects how VideoPhysEdit estimates physical parameters and simulates the edit rather than the off-the-shelf perception models it builds on.
Its videos show the full view of the source video at $720\times480$, and we resize them to the aspect ratio of the source video before scoring.

We score these videos by the protocol of Section~\ref{sec:supp_cf}, tracking their objects with the same trackers and checking every track by hand, and compute Mask IoU, PSNR, SSIM, and LPIPS as for the other models.
The results of the seven video editing models in Table~\ref{tab:videophysedit} of the main paper come from the videos that they generate for these 10 cases in the main evaluation.

\section{Additional Results}
\label{sec:supp_results}
This section breaks down the results of Section~\ref{sec:experiments} of the main paper by set, by object, by execution frame, by physical quantity, and by case.

\subsection{Simulated and Real Sets}
\label{sec:supp_results_split}
Table~\ref{tab:supp_split} gives the CF, EA, and Mask IoU of every model on the simulated and the real set, which Table~\ref{tab:main} of the main paper reports together.
On the real set, as on the whole benchmark, the commercial models score above the open-source models on CF and EA.

\begin{table}[!htbp]
\centering
\caption{\textbf{Results on the simulated and the real set.} Bold marks the column best.}
\label{tab:supp_split}
\small
\setlength{\tabcolsep}{5pt}
\begin{tabular*}{0.84\linewidth}{@{\extracolsep{\fill}}lcccccc}
\toprule
\multirow{2}{*}[-0.55ex]{\textbf{Model}} & \multicolumn{3}{c}{\textbf{Simulated}} & \multicolumn{3}{c}{\textbf{Real}} \\
\cmidrule(lr){2-4}\cmidrule(lr){5-7}
 & CF & EA & Mask IoU & CF & EA & Mask IoU \\
\midrule
\multicolumn{7}{l}{\textit{Open-source}} \\
JoyAI & 0.097 & 1.83 & 0.336 & 0.164 & 2.64 & 0.246 \\
Kiwi-Edit & $-$0.010 & 1.83 & 0.327 & 0.104 & 2.60 & \textbf{0.291} \\
OmniWeaving & $-$0.182 & 1.56 & 0.229 & $-$0.186 & 1.60 & 0.084 \\
\midrule
\multicolumn{7}{l}{\textit{Commercial}} \\
Kling 4.0 & 0.098 & 1.83 & \textbf{0.338} & \textbf{0.274} & 2.84 & 0.244 \\
MiniMax H3 & \textbf{0.131} & \textbf{2.16} & 0.311 & 0.177 & 2.96 & 0.224 \\
Seedance 2.5 & 0.120 & 2.02 & 0.337 & 0.242 & 3.12 & 0.205 \\
Wan 3.0 & 0.105 & 1.94 & 0.333 & 0.264 & \textbf{3.16} & 0.211 \\
\bottomrule
\end{tabular*}
\end{table}

\subsection{Removals by Object}
\label{sec:supp_results_removal}
Table~\ref{tab:supp_removal} splits the CF of the simulated removals between the removed object and the other objects, as the analysis of edit types in Section~\ref{sec:analysis} of the main paper reports.
For every video that performs a removal whose removed object has scored quantities, we average the scores on the removed object and those on the other objects separately, and then average each over the videos of a model.

\begin{table}[!htbp]
\centering
\caption{\textbf{CF of the simulated removals by object.}}
\label{tab:supp_removal}
\small
\setlength{\tabcolsep}{8pt}
\begin{tabular*}{0.58\linewidth}{@{\extracolsep{\fill}}lcc}
\toprule
\textbf{Model} & \textbf{Removed object} & \textbf{Other objects} \\
\midrule
\multicolumn{3}{l}{\textit{Open-source}} \\
JoyAI & 0.94 & $-$0.02 \\
Kiwi-Edit & 0.89 & $-$0.08 \\
OmniWeaving & 0.65 & $-$0.59 \\
VOID~\cite{Motamed2026VOID} & 0.92 & 0.08 \\
\midrule
\multicolumn{3}{l}{\textit{Commercial}} \\
Kling 4.0 & 0.89 & 0.02 \\
MiniMax H3 & 0.92 & 0.09 \\
Seedance 2.5 & 0.94 & $-$0.04 \\
Wan 3.0 & 0.93 & 0.08 \\
\bottomrule
\end{tabular*}
\end{table}

\subsection{Removals at a Later Frame}
\label{sec:supp_results_later}
Table~\ref{tab:supp_later} counts for every model the removals at a later execution frame $t$ in which the object is already gone before $t$, which Section~\ref{sec:analysis} of the main paper sums over the seven models.

\begin{table}[!htbp]
\centering
\caption{\textbf{Removals at a later frame.} Early counts, among the videos of the nine removals at a later execution frame $t$, those in which the removed object is absent in at least three frames before $t$.}
\label{tab:supp_later}
\small
\setlength{\tabcolsep}{16pt}
\begin{tabular}{lc}
\toprule
\textbf{Model} & \textbf{Early} \\
\midrule
\multicolumn{2}{l}{\textit{Open-source}} \\
JoyAI & 8 of 9 \\
Kiwi-Edit & 6 of 9 \\
OmniWeaving & 7 of 9 \\
VOID~\cite{Motamed2026VOID} & 8 of 9 \\
\midrule
\multicolumn{2}{l}{\textit{Commercial}} \\
Kling 4.0 & 6 of 9 \\
MiniMax H3 & 4 of 9 \\
Seedance 2.5 & 5 of 9 \\
Wan 3.0 & 8 of 9 \\
\midrule
Seven models & 44 of 63 \\
\bottomrule
\end{tabular}
\end{table}

\subsection{Physical Quantities}
\label{sec:supp_results_quantity}
Table~\ref{tab:supp_quantity} gives the mean CF and EA of every model on the changes of each physical quantity, which Section~\ref{sec:analysis} of the main paper summarizes by the EA in Figure~\ref{fig:quantity}, and Table~\ref{tab:breakdown} of the main paper gives the means over all changes.
We assign each change to the group of the quantity that it changes, with the wind speed in the gravity group and damping and drag in the friction group.

\begin{table}[!htbp]
\centering
\caption{\textbf{Results on the changes of each physical quantity.} Cases gives the number of changes of each quantity, and bold marks the column best of CF and of EA.}
\label{tab:supp_quantity}
\footnotesize
\setlength{\tabcolsep}{3.5pt}
\begin{tabularx}{\linewidth}{l*{8}{>{\centering\arraybackslash}X}}
\toprule
\textbf{Model} & \textbf{Flow rate} & \textbf{Density, viscosity} & \textbf{Stiffness} & \textbf{Friction, damping} & \textbf{Restitution} & \textbf{Mass} & \textbf{Velocity, position} & \textbf{Gravity, wind} \\
Cases & 24 & 32 & 46 & 45 & 22 & 84 & 65 & 16 \\
\midrule
\multicolumn{9}{c}{\textbf{CF}} \\
\multicolumn{9}{l}{\textit{Open-source}} \\
JoyAI & \textbf{0.109} & $-$0.016 & $-$0.010 & 0.006 & $-$0.012 & \textbf{0.009} & $-$0.016 & 0.035 \\
Kiwi-Edit & $-$0.231 & $-$0.175 & $-$0.168 & $-$0.067 & $-$0.174 & $-$0.144 & $-$0.060 & $-$0.196 \\
OmniWeaving & $-$0.210 & $-$0.116 & $-$0.180 & $-$0.073 & $-$0.459 & $-$0.303 & $-$0.359 & $-$0.165 \\
\midrule
\multicolumn{9}{l}{\textit{Commercial}} \\
Kling 4.0 & $-$0.021 & $-$0.111 & $-$0.005 & 0.038 & 0.013 & 0.003 & \textbf{0.015} & 0.007 \\
MiniMax H3 & 0.014 & \textbf{0.050} & \textbf{0.062} & $-$0.014 & \textbf{0.069} & $-$0.011 & $-$0.010 & \textbf{0.142} \\
Seedance 2.5 & 0.023 & $-$0.042 & 0.036 & 0.048 & $-$0.010 & $-$0.033 & 0.009 & 0.002 \\
Wan 3.0 & $-$0.096 & $-$0.055 & 0.020 & \textbf{0.053} & 0.030 & $-$0.077 & 0.014 & 0.007 \\
\midrule
\multicolumn{9}{c}{\textbf{EA}} \\
\multicolumn{9}{l}{\textit{Open-source}} \\
JoyAI & 1.67 & 1.72 & 1.61 & 1.44 & 1.18 & 1.29 & 1.18 & 1.00 \\
Kiwi-Edit & 1.83 & \textbf{1.91} & 1.54 & 1.38 & 1.00 & 1.29 & 1.28 & 1.12 \\
OmniWeaving & 1.42 & 1.72 & 1.41 & 1.38 & 1.18 & 1.13 & 1.09 & 1.38 \\
\midrule
\multicolumn{9}{l}{\textit{Commercial}} \\
Kling 4.0 & 1.58 & 1.72 & 1.43 & 1.40 & 1.18 & 1.26 & 1.23 & 1.06 \\
MiniMax H3 & 1.71 & 1.72 & \textbf{1.80} & \textbf{1.80} & \textbf{1.77} & \textbf{1.68} & \textbf{1.45} & \textbf{1.81} \\
Seedance 2.5 & 1.62 & 1.47 & 1.71 & 1.75 & 1.41 & 1.46 & 1.40 & \textbf{1.81} \\
Wan 3.0 & \textbf{2.17} & 1.47 & \textbf{1.80} & 1.38 & 1.09 & 1.24 & 1.20 & 1.06 \\
\bottomrule
\end{tabularx}
\end{table}

\subsection{Physics-Based Editing}
\label{sec:supp_results_vpe}
Table~\ref{tab:supp_vpe} gives the CF of the seven models and VideoPhysEdit on each of the 10 cases of Table~\ref{tab:videophysedit} of the main paper, with the means over the eight changes and the two removals.
VideoPhysEdit reaches the highest CF on seven of the eight changes, from which its advantage in Section~\ref{sec:analysis} of the main paper comes.

\begin{table}[!htbp]
\centering
\caption{\textbf{CF on each case with classic rigid-body motion.} Bold marks the best of each row.}
\label{tab:supp_vpe}
\footnotesize
\setlength{\tabcolsep}{3pt}
\resizebox{\linewidth}{!}{%
\begin{tabular}{>{\raggedright\arraybackslash}m{5.0cm}cccccccc}
\toprule
\multirow{2}{*}[-0.55ex]{\textbf{Edit}} & \multicolumn{3}{c}{\textbf{Open-source}} & \multicolumn{4}{c}{\textbf{Commercial}} & \multirow{2}{*}[-0.55ex]{VideoPhysEdit} \\
\cmidrule(lr){2-4}\cmidrule(lr){5-8}
 & JoyAI & Kiwi-Edit & OmniWeaving & Kling 4.0 & MiniMax H3 & Seedance 2.5 & Wan 3.0 & \\
\midrule
\multicolumn{9}{l}{\textit{Change}} \\
Reduce the restitution of the basketball to~0.5$\times$ & 0.008 & 0.025 & $-$1.000 & $-$0.055 & $-$0.091 & 0.090 & 0.073 & \textbf{0.990} \\
Reduce the initial velocity of the basketball to~0.3$\times$ & $-$0.051 & $-$0.012 & 0.062 & 0.000 & 0.057 & 0.000 & $-$0.030 & \textbf{0.436} \\
Increase the mass of the right soccer ball to~3$\times$ & 0.031 & $-$0.183 & $-$1.000 & 0.032 & $-$0.006 & \textbf{0.221} & 0.058 & $-$0.227 \\
Increase the initial velocity of the purple toy car to~3$\times$ & 0.001 & 0.045 & 0.034 & 0.016 & $-$0.100 & 0.000 & 0.005 & \textbf{0.518} \\
Increase the initial velocity of the toy car to~1.5$\times$ & $-$0.005 & 0.035 & $-$1.000 & $-$0.012 & 0.041 & 0.011 & 0.030 & \textbf{0.129} \\
Increase the mass of the left bowl to~5$\times$ & $-$0.017 & 0.092 & 0.012 & $-$0.056 & 0.007 & 0.040 & $-$0.001 & \textbf{0.687} \\
Increase the mass of the round tin to~5$\times$ & $-$0.129 & $-$0.488 & $-$0.719 & $-$0.037 & 0.003 & 0.344 & 0.056 & \textbf{0.496} \\
Increase the mass of the blue-handled stone to~3$\times$ & 0.022 & 0.042 & $-$0.580 & 0.190 & $-$0.018 & 0.000 & $-$0.031 & \textbf{0.429} \\
Mean & $-$0.018 & $-$0.056 & $-$0.524 & 0.010 & $-$0.014 & 0.088 & 0.020 & \textbf{0.432} \\
\midrule
\multicolumn{9}{l}{\textit{Remove}} \\
Remove the soccer ball & 0.499 & 0.492 & $-$0.004 & 0.491 & 0.747 & $-$0.001 & 0.487 & \textbf{0.881} \\
Remove the upright can & 0.406 & 0.425 & 0.433 & 0.423 & \textbf{0.808} & $-$0.100 & 0.432 & 0.469 \\
Mean & 0.453 & 0.458 & 0.215 & 0.457 & \textbf{0.778} & $-$0.051 & 0.460 & 0.675 \\
\midrule
Mean of all 10 & 0.076 & 0.047 & $-$0.376 & 0.099 & 0.145 & 0.060 & 0.108 & \textbf{0.481} \\
\bottomrule
\end{tabular}}
\end{table}
\FloatBarrier

\section{More Qualitative Examples}
\label{sec:supp_qualitative}
\begin{figure}[H]
\centering
\includegraphics[width=\linewidth]{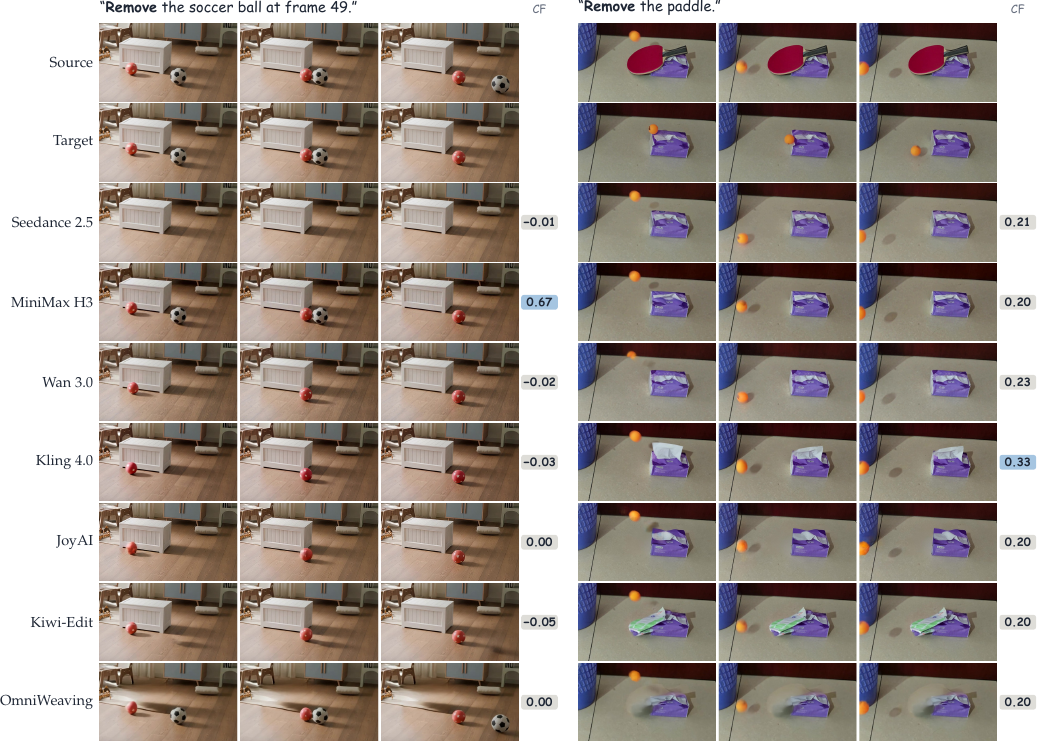}
\caption{\textbf{Removals.} Each block shows one case with the CF of each generated video, and the second case comes from the real set. In the target videos, the soccer ball disappears at frame 49, and the ping-pong ball hits the tissue packet instead of the paddle.}
\label{fig:supp_qual_removal}
\end{figure}

\begin{figure}[H]
\centering
\includegraphics[width=\linewidth]{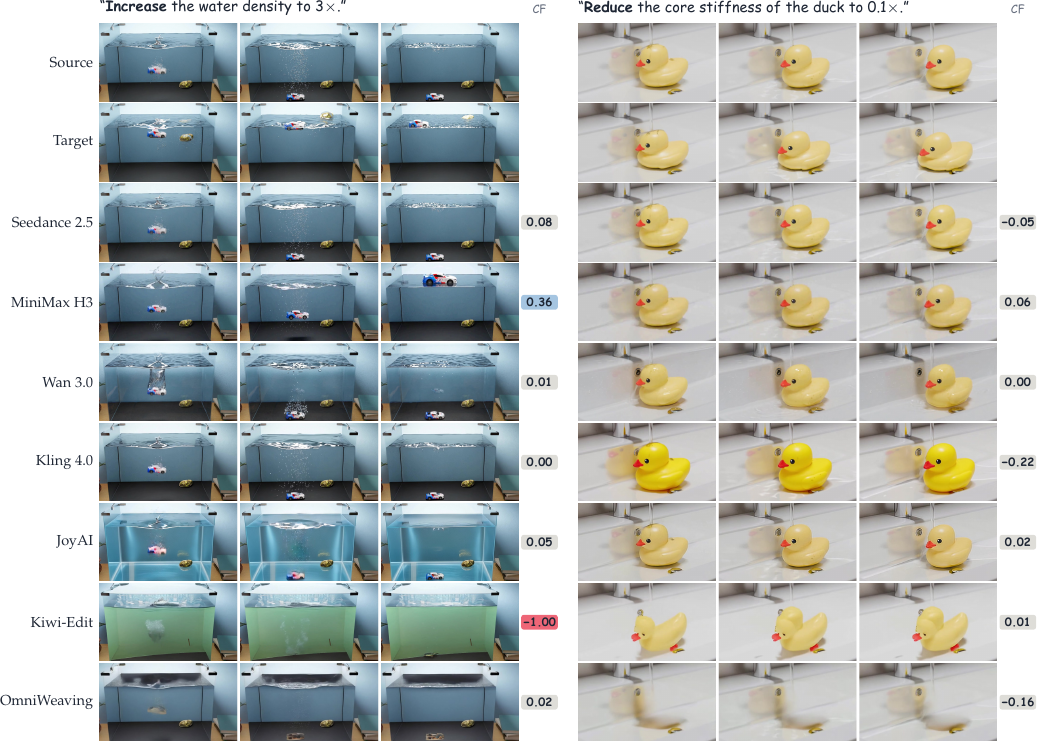}
\caption{\textbf{Changes of a physical quantity.} Each block shows one case with the CF of each generated video. In the target videos, the denser water lifts the toy car and the rock to the surface, and the softer core lets the duck sag under the water stream.}
\label{fig:supp_qual_change}
\end{figure}

\begin{figure}[H]
\centering
\includegraphics[width=\linewidth]{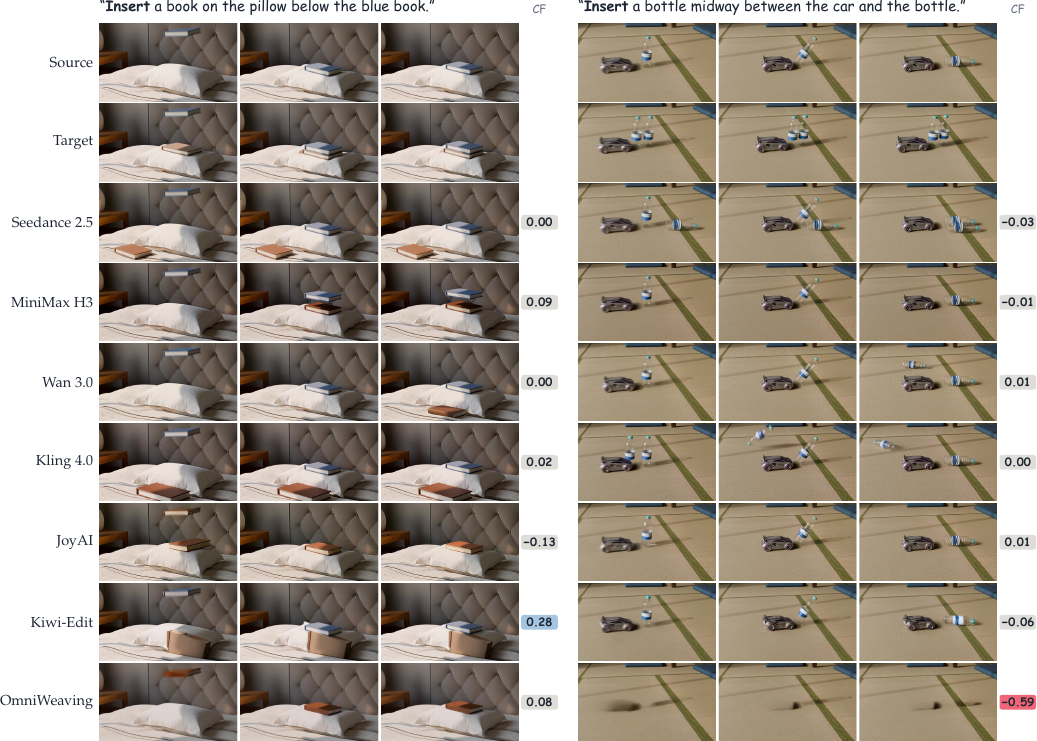}
\caption{\textbf{Insertions.} Each block shows one case with the CF of each generated video. In the target videos, the falling book lands on the inserted book, and the car stops in front of the two bottles, which stay upright.}
\label{fig:supp_qual_insertion}
\end{figure}

Six more cases illustrate that the models change how a scene looks but rarely how it evolves.
Figures~\ref{fig:supp_qual_removal}, \ref{fig:supp_qual_change}, and \ref{fig:supp_qual_insertion} show two removals, two changes of a physical quantity, and two insertions, with the CF of every generated video.

In removals, the models mostly remove only the appearance of an object, and a removal at a later frame often starts from the first frame.
The soccer ball should disappear at frame 49, but most models remove it from the first frame and obtain a CF near zero.
The models remove the paddle, but the ping-pong ball still follows its path from the source video and leaves the frame early.

\clearpage

Changes of a physical quantity are the edits that the models follow least, and some models change the look of the edited material instead.
In the denser water, most models leave the toy car on the bottom or lose it, and some tint the water instead, and none lifts both the car and the rock.
No model lets the duck sag when its core becomes softer, and some change only its color or its pose.

\clearpage

In insertions, the new object rarely acts on the other objects.
Most models place the new book or bottle off the path of the falling book or the car, add it late, recolor an existing object, or add nothing, so the book lands and the car knocks over the bottle as in the source video.

\end{document}